\documentclass[10pt,journal,compsoc]{IEEEtran}
\ifCLASSOPTIONcompsoc
  \usepackage[nocompress]{cite}
\else
  \usepackage{cite}
\fi
\ifCLASSINFOpdf
\else
 \usepackage[dvips]{graphicx}
\fi
\usepackage{algorithmic}
\usepackage{url}
\usepackage{amsmath,amsfonts}
\usepackage{algorithmic}
\usepackage{algorithm}
\usepackage{array}
\usepackage[caption=false,font=normalsize,labelfont=sf,textfont=sf]{subfig}
\usepackage{textcomp}
\usepackage{stfloats}
\usepackage{url}
\usepackage{verbatim}
\usepackage{graphicx}
\usepackage{cite}
\usepackage{multirow}
\usepackage{color}
\usepackage{amssymb}

\begin{document}
\title{Degradation Conditions Guided Cross-Consistent Deep Unfolding Network for All-In-One Video Restoration}

\author{Yuanshuo~Cheng, Mingwen Shao,~\IEEEmembership{Member,~IEEE,} Yecong~Wan, Yuanjian~Qiao, Wangmeng~Zuo,~\IEEEmembership{Senior~Member,~IEEE,} and~Deyu~Meng,~\IEEEmembership{Member,~IEEE}% <-this % stops a space
\IEEEcompsocitemizethanks{\IEEEcompsocthanksitem Yuanshuo Cheng, Mingwen Shao, Yecong Wan, and Yuanjian Qiao are with the College of Computer Science and Technology, China University of Petroleum, China.\protect\\
%E-mail: see http://www.michaelshell.org/contact.html
\IEEEcompsocthanksitem Wangmeng Zuo is with the School of Computer Science and Technology, Harbin Institute of Technology, China.\protect\\
\IEEEcompsocthanksitem Deyu Meng is with the Institute for Information and System Sciences, Xi’an Jiaotong University, China.}% <-this % stops an unwanted space
}

\markboth{Journal of \LaTeX\ Class Files,~Vol.~14, No.~8, August~2015}%
{Shell \MakeLowercase{\textit{et al.}}: Bare Demo of IEEEtran.cls for Computer Society Journals}

\IEEEtitleabstractindextext{%
\begin{abstract}
Existing Video Restoration (VR) methods always necessitate the individual deployment of models for each adverse weather degradations to remove them, lacking the capability for adaptive processing of degradations. Such limitation amplifies the complexity and deployment costs in practical applications. To overcome this deficiency, in this paper, we propose a Cross-consistent Deep Unfolding Network (CDUN) for All-In-One VR, which enables the employment of a single model to remove diverse degradations for the first time. Specifically, the proposed CDUN accomplishes a novel iterative optimization framework, capable of restoring frames corrupted by arbitrary degradations according to the degradation condations given in advance. To empower the framework for eliminating diverse degradations, we devise a Sequence-wise Adaptive Degradation Estimator (SADE) to estimate degradation condations for the corrupted video. By orchestrating these two cascading procedures, the proposed CDUN achieves a unified video restoration under diverse adverse weather degradations. In addition, we propose a window-based inter-frame fusion strategy to utilize information from more adjacent frames. This strategy involves the progressive stacking of temporal windows in multiple iterations, effectively enlarging the temporal receptive field and enabling each frame's restoration to leverage information from distant frames. Extensive experiments demonstrate that our method achieves state-of-the-art in All-In-One VR. Our code is available at https://github.com/YuanshuoCheng/CDUN.
\end{abstract}

}

\maketitle

\IEEEdisplaynontitleabstractindextext

\IEEEpeerreviewmaketitle

\IEEEraisesectionheading{\section{Introduction}\label{1}}

\IEEEPARstart{V}{ideos} captured in the real world often suffer from quality reduction due to adverse weather degradations such as rain, haze, snow, and low light. Consequently, the performances of subsequent downstream tasks such as object detection \cite{carion2020end,wang2023yolov7} and segmentation \cite{chen2018encoder,yu2022k} will be significantly impaired. In order to restore clean backgrounds from frames corrupted by degenerations, a number of Video Restoration (VR) methods have been proposed to utilize complementary redundancy information in adjacent frames to restore the damaged content. These methods can be categorized into explicit \cite{xue2019video,tian2020temporally,chan2021understanding} and implicit paradigms \cite{zhu2022deep,xia2020basis}. Explicit methods generally perform frame alignment among several adjacent frames based on optical flow \cite{dosovitskiy2015flownet} or deformable convolution \cite{dai2017deformable}, which subsequently fuses background features from multiple frames in an aligned feature space (in the context of an image or frame, the portion other than the degradation is referred to as the "background", "clean frame" or "clean image"). As for implicit methods, it is typical to train an end-to-end neural network to directly mine helpful information from unaligned adjacent frames. While existing efforts have made promising progress in video restoration, they all follow a "task-specific" paradigm. This means that a single model can only remove one specific type of degradation and cannot adaptively handle different degradations with a unified model. In real-world scenarios with diverse degradations, existing methods necessitate training and deploying models separately for each degradation type, which results in exorbitant application costs. Therefore, restoring videos corrupted by different degradations using a unified model (All-In-One VR) is a pressing issue.
\begin{figure}
	\begin{center}
		\includegraphics[width=\linewidth]{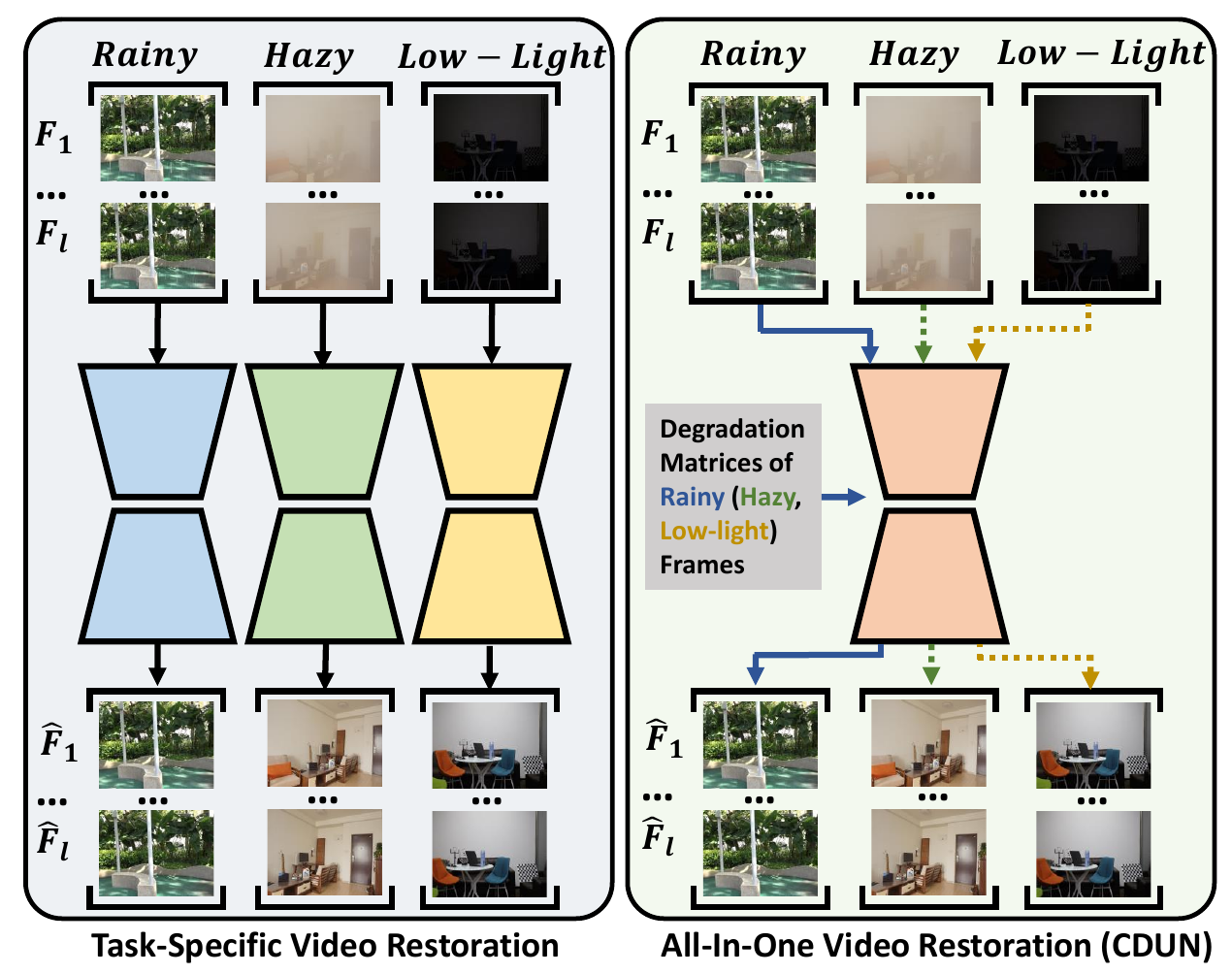}
	\end{center}
	\vspace{-15pt}
	\caption{Task-Specific VR require designing and training separate models for each type of degradation. The proposed CDUN achieves adaptive handling of different degradation types by adaptively adjusting degradation matrices, leading to the first-ever realization of an All-In-One VR.}
	\label{img_allinone}
	\vspace{-10pt}
\end{figure}

Most recently, several efforts have been made to learn to remove diverse unknown degradations from corrupted images using a single model \cite{li2020all,li2022all,chen2022learning,park2023all,zhu2023learning}. Although these methods achieve promising performances in Image Restoration (IR), they suffer from severe limitations in processing video data caused by their inability to exploit complementary redundant features among adjacent frames. Therefore, the All-In-One VR involves not only (1) enabling a single model to autonomously adapt to different degradation removal tasks, but also (2) realizing the extraction of complementary background information from multiple adjacent frames to improve the quality of the restored frames. The two requirements mentioned above define a task scenario with a high degree of intricacy, which leads to significant difficulties in modeling and solving the All-In-One VR problem.

Since the deep unfolding networks (DUNs) \cite{zhang2017learning,mou2022deep,wu2022uretinex} allow for flexible scenario modeling and several works propose using DUNs to tackle complex inverse problems with special requirements. For example, by extracting the general form of the super-resolution tasks with different blurring levels as a unified framework, the super-resolution under the effects of different blurring kernels can be solved in a unified manner \cite{zhang2020deep}. In addition, a related study \cite{sun2023deep} addresses the challenge of removing mixed noises in videos through sequential scene modeling. However, there is not yet an effective DUN-based method to deal with the adaptive removal of multi-degradation in videos. This is attributed to the fact that the features of different degradations are greatly different, making it difficult to extract a general form to serve as a unified fundamental framework. Moreover, modeling a long-range sequence of frames is a tough problem, which leads to the inability of existing DUN-based methods to capture information from frames with a longer temporal range.
%The Deep Unfolding Network (DUN) \cite{zhang2017learning,mou2022deep,wu2022uretinex} is a feasible paradigm to deal with the above challenges. Firstly, the DUNs can restore images corrupted by diverse degradations according to the pre-given features of different degradations \cite{zhang2020deep}. Secondly, the DUNs can enable extracting complementary features by modeling the correspondences among adjacent frames \cite{sun2023deep}. Nevertheless, such methods require providing features of each degradation in advance and are unable to adaptively generate them when dealing with different degradations. Therefore, they do not allow for the unified removal of diverse degradations. Besides, existing DUN-based methods are only capable of modeling inter-frame correlations within a short temporal window, which leads to a narrow temporal receptive field. This shortcoming limits the capturing of useful information from a longer temporal-range of frames.

To address the aforementioned issues, in this paper, we proposed a DUN-based All-In-One VR framework (as shown in Fig. \ref{img_allinone}), called Cross-Consistent Deep Unfolding Network (CDUN), to remove diverse adverse weather degradations in videos with a single model. The proposed CDUN implements adaptive estimation for unknown degradation features to guide the model to remove diverse degradations. And meanwhile, it effectively utilizes the temporal complementary information from long temporal-range frames.

Specifically, we propose a novel Sequence-wise Adaptive Degradation Estimator (SADE) to adaptively estimate the features of unknown degradation in the input video. Conditioned on the estimated degradation features, we construct a novel iterative optimization framework based on Maximum A Posteriori (MAP) to solve for the clean background corrupted by this degradation. 
The above two procedures are cascaded through the proposed cross-consistency strategy, which consequently realizes the autonomous adaptation to diverse degradation removal tasks. In addition, the cross-consistency strategy also align the background features of the frames in the same temporal window. This enables the restoration of each frame to obtain helpful information from multiple frames in the same temporal window.
%With the above two cascading procedures, the proposed method can realize autonomous adaptation for diverse degradation removal tasks. In addition, we propose a cross-consistency strategy to align the background features of the frames in the same temporal window. This strategy enables the restoration of each frame to obtain helpful information from multiple frames in the same temporal window. 
More importantly, the progressively iterative architecture of CDUN leads to the stacking of multiple temporal windows, thereby expanding the temporal receptive field. Thus, after multi-step iterative optimization, the proposed method can capture background information from frames at a greater distance rather than being limited to within only one temporal window.

In summary, the main contributions of this work are as follows:
%\begin{itemize}
%	\item In this paper, we propose an All-In-One VR framework called CDUN for the first time, which enables the unified removal of diverse adverse weather degradations from videos employing a single model. Extensive experiments demonstrate that the proposed CDUN outperforms existing methods. 
%	\item  We propose a novel SADE to estimate features of unknown degradations in the input videos, which subsequently guides an optimization framework to handle diverse degradation removal tasks according to the estimated features.
%	\item A stacked-window-based background feature fusion strategy is devised, which utilizes the iterative architecture to progressively expand the temporal receptive field. This enables the capturing and utilization of complementary background information from a long temporal-range of frames.
%	\item A video modeling strategy called cross-consistency is proposed, which integrates the aforementioned multi-degradation removal and long temporal range information fusion interpretably in a unified framework.
%\end{itemize}

\begin{itemize}
	\item  We devise a SADE to estimate features of unknown degradations in the input videos, which subsequently guides a novel unfolding-based optimization framework to handle diverse degradation removal tasks according to the estimated features.
	\item A stacked-window-based background feature fusion strategy is developed, which utilizes the iterative architecture to progressively expand the temporal receptive field. This enables the capturing and utilization of complementary background information from a long temporal-range of frames.
	\item We propose a novel cross-consistency scheme for video modeling in multi-degradation scenarios. It integrates the aforementioned multi-degradation removal and long-temporal-range information fusion interpretably in a unified framework. 
	\item Our cross-consistency-based CDUN is the first work to realize All-In-One VR. Qualitative and quantitative experiments on multiple datasets demonstrate that our method outperforms existing methods in All-In-One video restoration.
\end{itemize}

\section{Related work}
\label{2}
\subsection{Video Restoration}
\label{2.1}
The VR task is dedicated to mining complementary background information from adjacent frames to restore the corrupted backgrounds. Most of the early works explicitly aligns adjacent frames based on the estimated optical flows \cite{dosovitskiy2015flownet,shi2023videoflow,pan2020cascaded} or the offsets of the deformable convolutions \cite{dai2017deformable,wang2019edvr}, which consequently enables the sharing and utilization of complementary features among these frames. Nevertheless, the accuracy of explicit alignment is susceptible to long displacements \cite{li2021arvo} and motion blur \cite{son2021recurrent}. Therefore, some subsequent works propose more robust implicit methods to utilize complementary features directly. For instance, Zhu et al.\cite{zhu2022deep} cancel the frame alignment procedure and focus solely on fusing features from unaligned adjacent frames via an well-designed neural network. In addition, limited by the computational cost, these methods are mostly constructed based on sliding window \cite{isobe2020video,li2021arvo,li2020mucan} or recurrent architectures \cite{chan2021basicvsr,isobe2020revisiting,nah2019recurrent}. Both of these architectures have a short temporal receptive field and are unable to utilize the information from more distant frames. To address this issue, Liang et al.\cite{liang2022vrt} propose a model that incorporates Temporal Mutual Self Attention and Parallel Warping to achieve long-range modeling capability. Although existing efforts have made promising progress in VR, a single model in these methods can only deal with one type of degradation. Therefore, the deployment of multiple models is required to cater to multi-degradation scenarios, which bear a high application cost. In contrast, our method is able to adaptively restore videos corrupted by diverse degradations with a single model.
\subsection{All-in-one Image Restoration}
\label{2.2}
Most recently, several efforts attempt to develop All-In-One IR \cite{li2020all,li2022all,wan2023restoring,zhang2023ingredient,cheng2023deep} frameworks to restore images corrupted by different degradations with a single model. 
%They overcome the high application cost problem in Task-Specific IR \cite{zamir2021multi,zamir2022restormer,wang2022uformer,wan2022image}, which is caused by the need to deploy multiple models for eliminating multiple degradations. 
The key to All-In-One IR lies in developing effective techniques to adapt a single model to different degradation types. To this end, Li et al.\cite{li2020all} devise a model with multiple encoders to remove multiple degradations, which is the first successful implementation of All-In-One IR. However, in order to choose the corresponding encoder, this approach requires providing the degradation type in advance, which hinders the implementation of an end-to-end framework. Therefore, some further works tend to develop more convenient frameworks to enable the model to autonomously discriminate the degradation type and remove the corresponding degradations. For instance, Chen \emph{et al.}\cite{chen2022learning} propose using knowledge distillation to enable an All-In-One student model to learn the removal capabilities of diverse degradations from multiple Task-Specific teacher models \cite{zamir2022restormer,wang2022uformer}. Zhu et al. \cite{zhu2023learning} further discover that different weather degradations exhibit specific and general features, and suggest enhancing the multi-degradation removal capability by learning these two features. Compared to these All-In-One IR methods, our proposed All-In-One VR framework not only can adaptively remove diverse degradations, but also utilizes the abundant complementary features in multiple frames to improve the quality of the restored frames.
\subsection{Deep Unfolding Network}
\label{2.3}
The IR task can be formulated as a MAP problem, which can be solved by minimizing the established energy function. To this end, some works introduce deep neural networks (DNNs) into traditional optimization algorithms to develop efficient solving methods, which is the fundamental paradigm of DUNs. Specifically, the deep unfolding methods unfold the energy function into a fidelity term-related subproblem and a regular term-related subproblem \cite{geman1995nonlinear,boyd2011distributed}, and minimize the energy function by alternately optimizing the two subproblems. The earlier works employ handcrafted prior constraints \cite{barbu2009training,samuel2009learning} to optimize the regular term-related subproblems, but they typically suffer from cumbersome crafting processes and a limited representation ability. To overcome this drawback, Zhang et al.\cite{zhang2017learning} firstly propose to use DNN prior to replace the hand-crafted prior, which simplifies the algorithmic process through end-to-end training. Meanwhile, DNNs can fit complex priors in a data-driven manner, which enables DUNs to effectively handle more challenging degradations such as low light \cite{wu2022uretinex}, shadow \cite{guo2023shadowdiffusion}, rain \cite{mou2022deep}, etc. In addition, the degradation model in the fidelity term determines the type of degradation that the DUNs can handle. Therefore, the DUNs can be adapted to different IR tasks by utilizing different degradation models. Leveraging this capability, Zhang et al.\cite{zhang2020deep} devised a DUN-based method that achieves unified image super-resolution with different noise levels and blur kernels. However, since the degradation features (i.e., noise and blur kernels) need to be given manually, this approach cannot implement an end-to-end framework. Unlike the aforementioned works, our method can flexibly restore videos corrupted by different degradations by adaptively adjusting the degradation model. %Meanwhile, it can also utilize the abundant complementary information in long temporal-range frames to restore the corrupted content.

\section{Methodology}
\label{3}
In this section, we present the overall framework of the proposed CDUN (see Fig. \ref{img_framework}, Algorithm \ref{al_cdun} and Sec. \ref{3.4}) and describe its detailed techniques. Specifically, we first introduced the proposed cross-consistency, which serves as the foundation for constructing CDUN. Based on cross-consistency, we provided the formalized representation of CDUN. Then, we presented the unfolding-based optimization strategy. Subsequently, we described the overall framework and training process of CDUN. Finally, we explained the structure and roles of the crucial modules.
\begin{figure}
	\begin{center}
		\includegraphics[width=\linewidth]{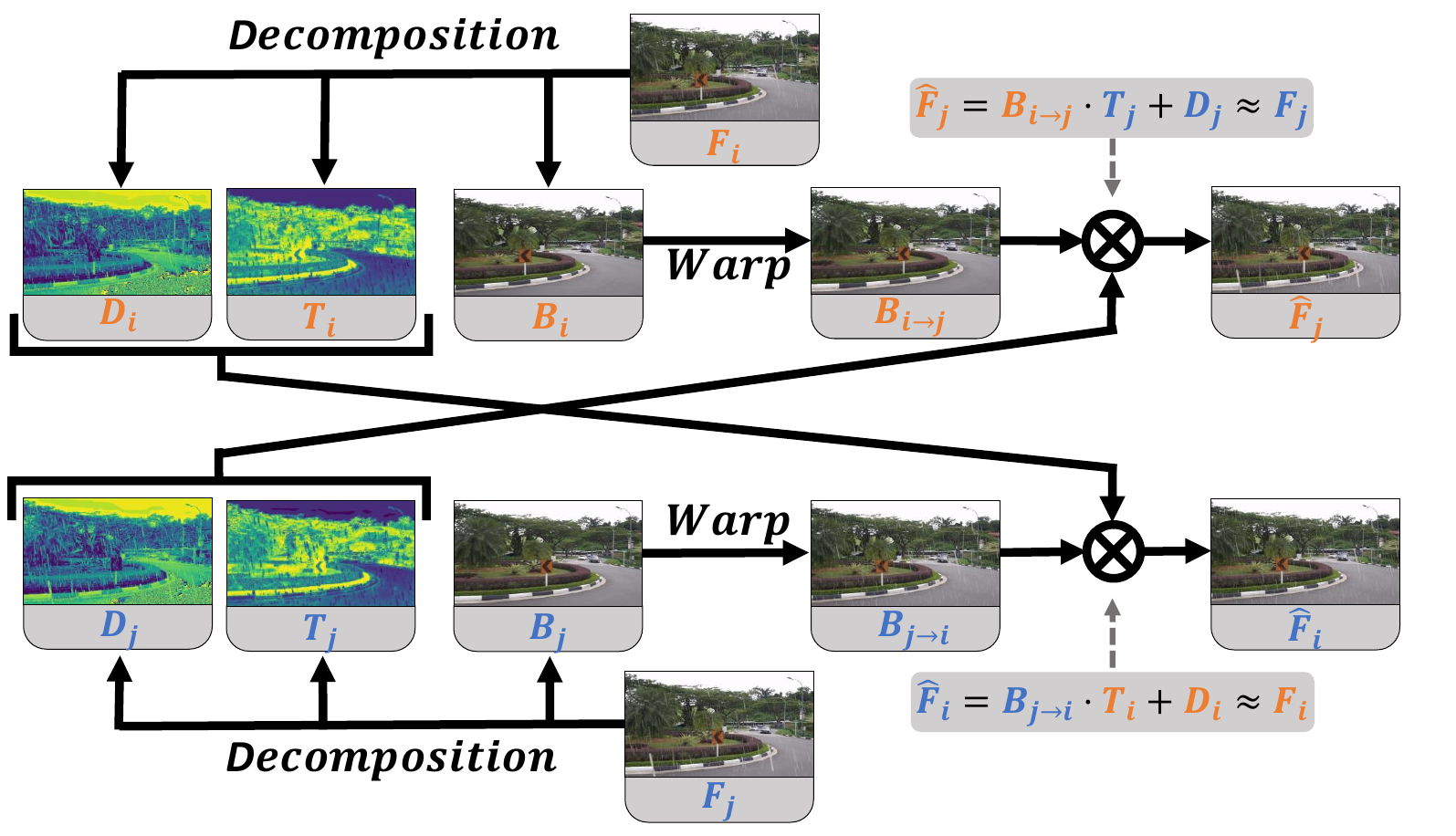}
	\end{center}
	\vspace{-15pt}
	\caption{Visualization of the cross-consistency. cross-consistency can be summarized as: For two adjacent frames (i.e., $F_i$, $F_j$), apply a motion operation to their background and exchange the degradation matrices. Consequently, they should transform into each other (i.e., $F_j$, $F_i$).}
	\label{img_cc}
\end{figure}
\subsection{Cross-Consistency}
\label{3.1}
Cross-consistency models the correlations among degraded frames, degradation matrices and clean frames within a temporal window. It guides the algorithmic model to obtain information from neighboring frames on the one hand, and on the other hand, provides a way to change the type of degradation that can be handled by the algorithmic model (i.e., change the degradation matrices).

We observed that most degradation types, such as rain\cite{huang2012context}, haze\cite{he2010single}, snow\cite{liu2018desnownet}, and low-light\cite{liu2021retinex}, can be modeled by an additive factor and a multiplicative factor (unless otherwise specified, all multiplication operations refer to element-wise multiplication):
\begin{equation}
F_i=T_i\cdot B_i+D_i, \label{eq_allinone}
\end{equation}
where $T_i$ and $D_i$ are collectively referred to as degradation matrices. $F_i$ represents the i-th frame in the frame sequence and $B_i$ represents the corresponding clean background of $F_i$, which is our desired objective for restoration. The adjustment of the degradation matrices makes it possible to model different types of degradation without changing the structure of the model, which is necessary for deve1loping an unified multitasking framework. For two adjacent degraded frames' clean backgrounds, $B_i$ and $B_j$, the pixels in them exhibit a certain motion relationship, which can be represented as $M_{i\rightarrow j}(B_i) \approx B_j$. $M_{i\rightarrow j}(*)$ is a motion (warping) operator for performing a motion operation on image $*$ according to the movement relation (optical flow) from $B_i$ to $B_j$ (see Sec. \ref{3.6} for implementation details). In this case, when we apply the degradation matrices of $F_j$ (i.e., $T_j$ and $D_j$) to $ M_{i\rightarrow j}(B_i)$, the resulting image should be approximately equal to $F_j$. Therefore, cross-consistency can be formulated as follows:
\begin{equation}
E_{B_i}[M_{i\rightarrow j}(B_i)\cdot T_j+D_j]=F_j, \label{eq_cc}
\end{equation}
where $E$ denotes the mathematical expectation. Additionally, Fig. \ref{img_cc} further visualizes the principles of cross-consistency.

\subsection{Problem Formulation}
\label{3.2}
Unlike IR, VR requires gathering information from adjacent frames to aid in the restoration process. Therefore, the MAP framework for video restoration can be represented as follows:
\begin{equation}
\begin{aligned}
\hat{B_i} &= \arg\mathop{\max}_{B_i}\sum_{j\in \mathcal{I}}\log P(B_i|F_j)\\
&=\arg\mathop{\max}_{B_i}[\frac{1}{2n+1}\sum_{j\in \mathcal{I}}\log P(F_j|B_i)+\log P(B_i)],
\end{aligned}
\label{eq_mapv}
\end{equation}
where $\hat{B_i}$ is our desired optimization objective, i.e., clean background. $\mathcal{I}$ represents a temporal window, and $\mathcal{I} = \{i-n, ..., i-1, i, i+1, ..., i+n\}$. In practice, $n$ is set to $1$, meaning that $\mathcal{I}$ is a window of length $3$. Based on the cross-consistency principle and the above-mentioned MAP framework, we represent the All-In-One VR as a problem of minimizing the following energy function:
\begin{equation}
\begin{aligned}
\hat{B_i}=\arg\mathop{\min}_{B_i}&\frac{1}{2n+1}\sum_{j\in \mathcal{I}}\frac{1}{2}||W_{i\rightarrow j}\{F_{j}-[M_{i\rightarrow j}(B_i)\\&\cdot T_j+D_j]\} ||^2+\lambda \Phi(B_i),
\label{eq_energyv}
\end{aligned}
\end{equation}
where $W_{i\rightarrow j}$ represents spatial motion weight, which is dependent on the motion distances of individual pixels. $W_{i\rightarrow j}$ can alleviate the issue of motion error interference during the restoration process. $\lambda$ is a trade-off parameter. All operations involving a single scalar and a matrix are equivalent to performing the operation on each element of the matrix with the scalar

As can be seen from the Eq. (\ref{eq_energyv}), the algorithm model can be adapted to different degradation removal tasks by adjusting the $T_j$ and $D_j$ for different degradations. %Here we summarize how the CDUN achieves adaptive handling of different degradations as : (1) decoupling the degradation matrices, degraded frames, and clean frames, through cross-consistency; (2)adaptively estimating the degradation matrices of degraded frames to enable the model to handle corresponding types of degradations.

\begin{figure*}
	\begin{center}
		\includegraphics[width=\linewidth]{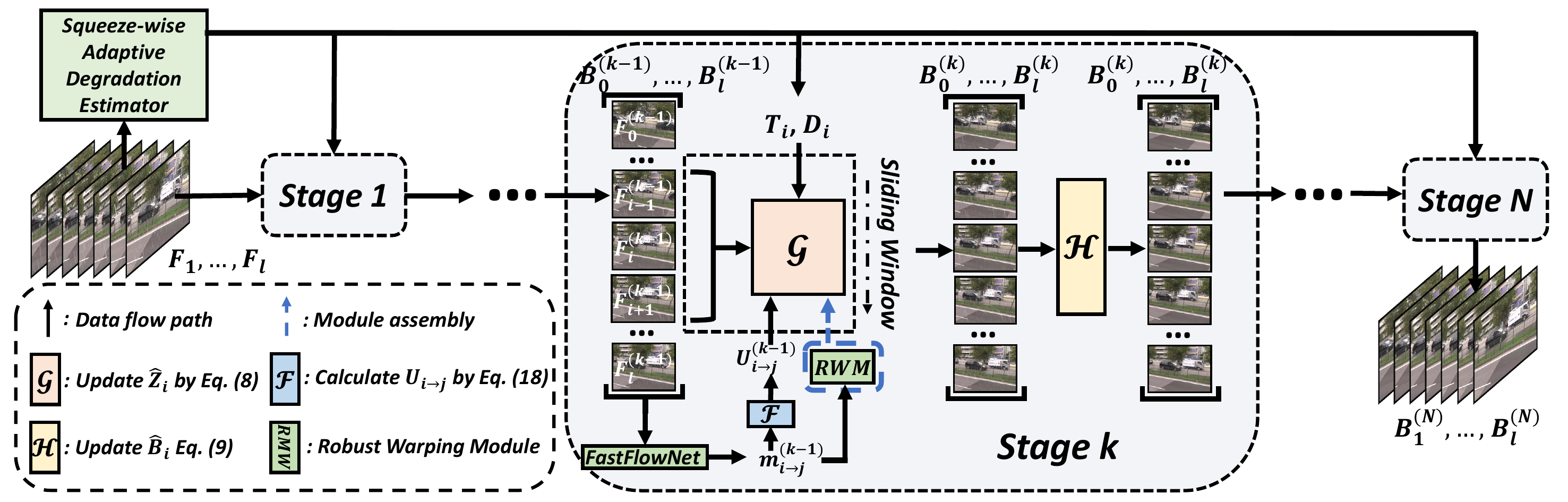}
	\end{center}
	\vspace{-15pt}
	\caption{Overall framework of Cross-Consistent Deep Unfolding Network (CDUN). The proposed CDUN achieves All-In-One VR for the first time. It addresses the challenges of adaptive handling of multiple degradations in VR and the enlargement of temporal receptive fields.}
	\label{img_framework}
\end{figure*}
\subsection{Unfolding Optimization}
\label{3.3}
By employing the Half Quadratic Splitting (HQS) algorithm\cite{geman1995nonlinear}, Eq. (\ref{eq_energyv}) can be unfolded into two sub-problems: one related to the fidelity term (Eq. (\ref{eq_sul3a})) and the other related to the regularization term (Eq. (\ref{eq_sul3b})). The optimization of Eq. (\ref{eq_energyv}) can then be achieved by iteratively optimizing these two sub-problems alternately. 
%The degradation matrices operate during the solution process of the fidelity term-related sub-problem. By adaptively estimating the degradation matrices, CDUN achieves control over the types of degradation being addressed. The optimization of the regularization term-related sub-problem is achieved by introducing a CNN prior. Compared to traditional Gaussian denoising priors\cite{zhang2017learning,zhang2017beyond}, the CNN prior offers the advantages of stronger robustness and convenience of acquisition through end-to-end training. 
Specifically, by introducing an auxiliary variable $Z$, Eq. (\ref{eq_energyv}) can be rewritten as follows:
\begin{equation}
\begin{aligned}
\hat{B_i}=\arg\mathop{\min}_{B_i}&\frac{1}{2n+1}\sum_{j\in \mathcal{I}}\frac{1}{2}||W_{i\rightarrow j}\{F_{j}\!-\![M_{i\rightarrow j}(Z_i)\\
&\cdot T_j\!+\!D_j]\} ||^2\!+\!\lambda \Phi(B_i) \text{ s.t. } Zi=Bi.
\label{eq_sul1}
\end{aligned}
\end{equation}
To handle the equality constraint in the above equation, we introduce a quadratic term and rewrite the equation in an unconstrained form:
\begin{equation}
\begin{aligned}
\hat{B_i}=\arg\mathop{\min}_{B_i}&\frac{1}{2n+1}\sum_{j\in \mathcal{I}}\frac{1}{2}||W_{i\rightarrow j}\{F_{j}-\![M_{i\rightarrow j}\!(\!Z_i\!)\\
&\cdot\! T_j\!+\!D_j]\} ||^2\!+\!\frac{\gamma}{2}\!||Z_i\!-\!B_i||^2\!+\!\lambda\Phi(\!B_i\!),
\label{eq_sul2}
\end{aligned}
\end{equation}
where $\gamma$ is a penalty parameter. According to the HQS algorithm\cite{geman1995nonlinear}, Eq. (\ref{eq_sul2}) can be decomposed into two sub-problems and optimized alternately:
%\begin{equation}
%\begin{aligned}
%\hat{Z_i}&=\arg\mathop{\min}_{Z_i}\frac{1}{2n+1}\sum_{j\in \mathcal{I}}\frac{1}{2}||W_{i\rightarrow j}\{F_{j}\\
%&-[M_{i\rightarrow j}(Z_i)\cdot T_j+D_j]\} ||^2+\frac{\gamma}{2}||Z_i-B_i||^2\\
%\hat{B_i}&=\arg\mathop{\min}_{B_i}\frac{\gamma}{2}||Z_i-B_i||^2+\lambda\Phi(B_i)
%\end{aligned}
%\end{equation}
\begin{subequations}
	\begin{align}
	\hat{Z_i}=\arg\mathop{\min}_{Z_i}&\frac{1}{2n+1}\sum_{j\in \mathcal{I}}\frac{1}{2}||W_{i\rightarrow j}\{F_{j}-[M_{i\rightarrow j}(Z_i) \nonumber\\
	&\cdot T_j+D_j]\} ||^2+\frac{\gamma}{2}||Z_i-B_i||^2, \label{eq_sul3a}\\
	\hat{B_i}=\arg\mathop{\min}_{B_i}&\frac{\gamma}{2}||Z_i-B_i||^2+\lambda\Phi(B_i). \label{eq_sul3b}
	\end{align}
\end{subequations}

Eq. (\ref{eq_sul3a}) is a least-squares problem, and we can directly calculate its closed-form solution (the detailed derivation and analysis is given in the supplementary material.):
\begin{equation}
\begin{aligned}
\hat{Z_i}=\frac{\gamma B_i\!+\!\sum_{j \in \mathcal{I}}U_{i\rightarrow j}M_{j\rightarrow i}[T_j(F_j\!-\!D_j)]/(2n\!+\!1)}{\sum_{j \in\mathcal{I}}U_{i\rightarrow j}M_{j\rightarrow i}(T_j^2)/(2n+1)+\gamma},\label{eq_updatez}
\end{aligned}
\end{equation}
where $U_{i\rightarrow j} = M_{j\rightarrow i}(W_{i\rightarrow j}^2)$. 
The update equation for $Z_i$ (i.e. (Eq. \ref{eq_updatez})) obtained based on cross-consistency has two crucial properties: (1) it provides an interface to control the type of degradation, i.e., $T_j$ and $D_j$. By estimating the degradation matrices of different degradations using the proposed SADE (see Sec. \ref{3.5}), the algorithmic model (i.e. Eq. (\ref{eq_energyv})) makes it possible to restore videos corrupted by these degradations in a unified framework; (2) it implements information interaction among the frames in a temporal window. The multi-step iterative process progressively expands the temporal receptive field, which consequently enables more frames to interact with each other and capture rich complementary information.
In practice, we directly compute $U_{i\rightarrow j}$ instead of $W_{i\rightarrow j}$ to simplify the calculations. The reason and related analysis is detailed in Sec. \ref{3.7}. The motion operator $M_{j\rightarrow i}$ is implemented through the proposed RWM, and the detailed implementation method is presented in Sec. \ref{3.6}. %$U_{i\rightarrow j}$ is related to the motion distances from $B_i$ to $B_j$, and the detailed calculation method is elaborated in Sec. \ref{3.7}.

For the optimization of the regularization term-related sub-problem (i.e., Eq. (\ref{eq_sul3b})), we trained a DNN model based on the U-Net\cite{ronneberger2015u} architecture to approximate the implicit prior. This process can be formulated as follows:
\begin{equation}
\begin{aligned}
\hat{B_i} = UNet(Z_i).\label{eq_updateb}
\end{aligned}
\end{equation}
%Compared to traditional Gaussian denoising priors\cite{zhang2017learning,zhang2017beyond}, the CNN prior offers the advantages of stronger robustness and convenience of acquisition through end-to-end training.
In practice, the parameters of the U-Net are not shared across different iterations. Compared to manually crafted priors (\emph{e.g.}, Gaussian denoising prior \cite{zhang2017learning,zhang2017beyond}), the DNN prior provides the advantages of stronger robustness and convenience of acquisition through end-to-end training.

\subsection{Overall Framework and Training Process for Cross-Consistent Deep Unfolding Network}
\label{3.4}
\begin{algorithm}[t]
	\caption{Cross-Consistent Deep Unfolding Network}
	\begin{algorithmic}[1]
		\STATE \textbf{Input:}$\mathcal{V}=\{F_i \in \mathbb{R}^{3 \times H \times W}|i=1,2,…l\}$\\ \hfill/* degraded video of length $l$ */
		\STATE \textbf{Output:}$\mathcal{B}^{(N)}=\{B_i^{(N)} \in \mathbb{R}^{3 \times H \times W}|i=1,2,…l\}$ \\
		\hfill/* clean video of length $l$ */
		\STATE $\{T_i,D_i \in \mathbb{R}^{1 \times H \times W}|i=1,2,…l\}=SADE({\mathcal{V}})$ \\
		\hfill/* estimate degradation matrice with Eq. (\ref{eq_sade}) */
		\STATE$\mathcal{B}^{(0)}=\{B_i^{(0)}|B_i^{(0)}=F_i,i=1,2,…l\}$
		\STATE $\gamma^0=0.1$
		\STATE $s=0.05$
		\hfill/* initialize $\mathcal{B}^{(0)}$, $\gamma^{(0)}$ and $s$*/
		\STATE \textbf{for} $k = 1 \to N$\textbf{:} \hfill/* $N$ iterations */
		\STATE \hspace{0.5cm}\textbf{for} $i = 1 \to l$\textbf{:}
		\STATE \hspace{1cm}\textbf{if} $i=1$\textbf{:}
		\STATE \hspace{1.5cm} $\mathcal{I} = \{2,1,3\}$
		\STATE \hspace{1cm}\textbf{else if} $i=l$\textbf{:}
		\STATE \hspace{1.5cm} $\mathcal{I} = \{l-2,l,l-1\}$
		\STATE \hspace{1cm}\textbf{else:}
		\STATE \hspace{1.5cm} $\mathcal{I} = \{i-1,i,i+1\}$ \hfill/* temporal window */
		\STATE \hspace{1cm}\textbf{end if}
		\STATE \hspace{1cm}\textbf{for} $j$ in $\mathcal{I}$\textbf{:}
		\STATE \hspace{1.5cm}$m^{(k-1)}_{i\rightarrow j} \!\in\! \mathbb{R}^{2 \times H \times W} \!=\!FastFlowNet(F^{(k-1)}_i  $,\\
		\hspace{4.5cm}$F^{(k-1)}_j)$ \hfill/*optical flow */
		\STATE \hspace{1.5cm}$M^{(k-1)}_{i\rightarrow j}(*)=RWM(*,m^{(k-1)}_{i\rightarrow j})$ \\ \hfill/* Eq. (\ref{eq_Ma}) and  Eq. (\ref{eq_Mb}), omitting objective matrix */
		\STATE \hspace{1.5cm}$U^{(k-1)}_{i\rightarrow j} \!\in\! \mathbb{R}^{1 \times H \times W} \!=\! \mathcal{F}(||m^{(k-1)}_{i\rightarrow j}||^2)$ \\
		\hspace{2cm} \hfill/* Eq. (\ref{eq_getU}) */
		\STATE \hspace{1cm}\textbf{end for}
		\STATE \hspace{1cm}$Z_i^{(k)}\!=\!\frac{\gamma B_i^{(k-1)}\!+\!\sum_{j \in \mathcal{I}}U^{(k-1)}_{i\rightarrow j}M^{(k-1)}_{j\rightarrow i}[T_j(F_j\!-\!D_j)]/(2n\!+\!1)}{\sum_{j \in\mathcal{I}}U^{(k-1)}_{i\rightarrow j}M^{(k-1)}_{j\rightarrow i}(T_j^2)/(2n+1)+\gamma^{(k-1)}}$\\
		\hspace{1.5cm} \hfill/* Eq. (\ref{eq_updatez}) */
		\STATE \hspace{1cm}$B^{(k)}_i=UNet(Z_i^{(k)})$ \hfill/* Eq. (\ref{eq_updateb}) */
		\STATE \hspace{0.5cm}\textbf{end for}
		\STATE\hspace{0.5cm}$\mathcal{B}^{(k)}=\{B_i^{(k)}|i=1,2,…l\}$
		\STATE\hspace{0.5cm}$\gamma^{(k)}=\gamma^{(k-1)}+s$
		\STATE \textbf{end for}
		\STATE Return $\mathcal{B}^{(N)}$
	\end{algorithmic}
	\label{al_cdun}
\end{algorithm}
%The proposed CDUN achieves excellent All-In-One video restoration performance by (1) adaptive handling of different types of degradations and (2) enabling longer temporal receptive fields.
Fig. \ref{img_framework} illustrates the overall framework of CDUN. First, we utilize the proposed SADE (see Sec. \ref{3.5} for details) to estimate the degradation matrices of the input degraded frames $\mathcal{V}=\{F_i \in \mathbb{R}^{3 \times H \times W}|i=1,2,…l\}$. Subsequently, we solve for the clean frames by iteratively optimizing Eq. (\ref{eq_energyv}). Since the motion operator $M_{i\rightarrow j}$ and the spatial motion weight are both based on optical flow, at the beginning of each iteration stage, we first estimate the optical flow among adjacent frames using FastFlowNet\cite{kong2021fastflownet}. Then, we iteratively execute Eq. (\ref{eq_updatez}) and Eq. (\ref{eq_updateb}) alternately to optimize Eq. (\ref{eq_energyv}). Similar to existing methods\cite{wu2022uretinex,zhang2020deep}, the penalty parameters $\gamma$ increases during the iterative process. The output sequence $\mathcal{B}^{(N)}=\{B_i^{(N)} \in \mathbb{R}^{3 \times H \times W}|i=1,2,…l\}$, obtained after $N$ iterations, represents the restored clean frames. We present the detailed pseudocode for the inference process of CDUN in Algorithm \ref{al_cdun}.

\begin{figure}
	\begin{center}
		\includegraphics[width=\linewidth]{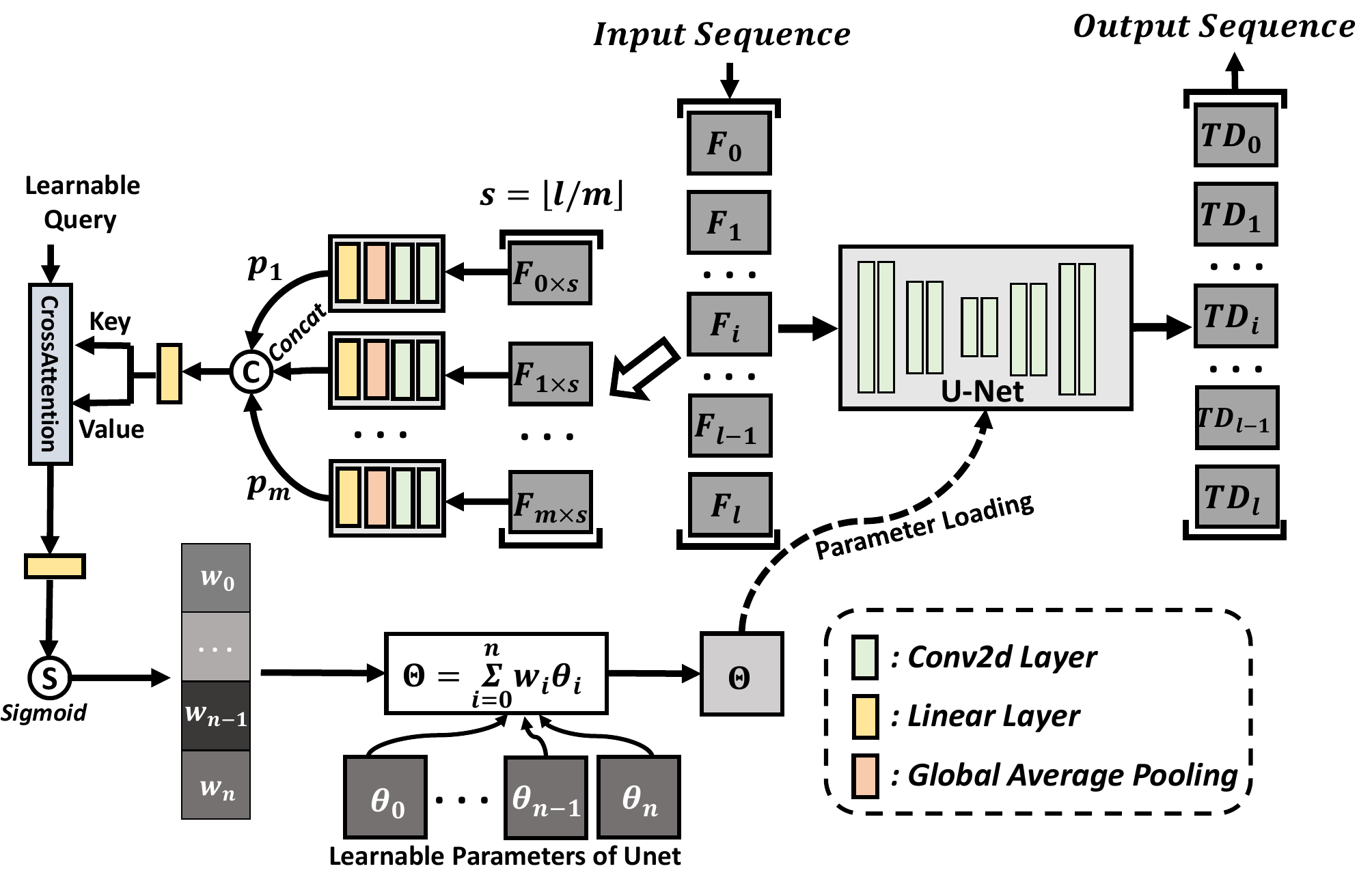}
	\end{center}
	\vspace{-15pt}
	\caption{Sequence-wise Adaptive Degradation Estimator (SADE). This work achieves adaptive processing for different degradation types based on adaptively estimating the degradation matrices using SADE.}
	\label{img_ade}
\end{figure}

As can be seen from the aforementioned procedures, the proposed method removes the degradations according to the corresponding degradation matrices. The algorithmic model is endowed with multi-degradation handling capability by the SADE that adaptively estimates various degradation matrices as conditional guidances. Moreover, Eq. (\ref{eq_updatez}) involves a sliding window process. Multiple iterations achieve the stacking of windows, thereby expanding the temporal receptive field.

The proposed method consists of a two-stage training process. In the first stage, we train the SADE in an All-In-One manner. The training details and the loss function can be found in Sec \ref{3.5}. Additionally, the parameters of FastFlowNet\cite{kong2021fastflownet} are provided by the official source and do not require additional training. Afterwards, we freeze the parameters of SADE and FastFlowNet and train CDUN in an end-to-end manner. The loss function $\mathcal{L}_{CDUN}$ consists of SSIM loss\cite{wang1987ssim} and Charbonnier loss\cite{charbonnier1994two} and can be represented as follows:
\begin{subequations}
	\begin{align}
	\mathcal{L}_{sup}(x,y)=&[1-SSIM(x,y)]\nonumber\\&+\sqrt{||x-y||^2+\epsilon^2},\label{eq_losssup}\\
	\mathcal{L}_{CDUN}(\mathcal{B}^{(N)},\mathcal{Y}))=&\frac{1}{l}\sum_{i=1}^l\mathcal{L}_{sup}(B_i^{(N)},Y_i),
	\label{eq_losscdun}
	\end{align}
\end{subequations}
where $\epsilon=0.001$. $\mathcal{Y}=\{Y_i \in \mathbb{R}^{3 \times H \times W}|i=1,2,…,l\}$ represents the paired grand-truth video.
\subsection{Sequence-wise Adaptive Degradation Estimator}
\label{3.5}
The SADE guides the algorithm model (i.e. Eq. (\ref{eq_energyv})) to adaptively restore videos corrupted by different degradations by adaptively estimating the degradation matrices of the input video.

Conditional networks (or dynamic networks) can adapt to different types of input data by adjusting the parameters \cite{shan2020meta,denil2013predicting} or structures \cite{park2015big, huang2017multi} of the model, which are applicable to process various kinds of degradation. Among them, parameter adaptive adjustment is a convenient and effective paradigm. Existing methods typically generate a set of parameters based on the input image's features and then process that image (referred to as image-wise adaptive or frame-wise adaptive). Since the degradation type in the frames of a same video clip is generally consistent, it is unnecessary to adjust the parameters once for each frame. The classical frame-wise adaptive paradigm not only wastes computational resources, but also suffers from instability when processing neighboring frames (see Sec. \ref{4.4} and Fig. \ref{img_dyw}). Therefore, we propose a sequence-wise adaptive strategy to deal with the this problems. Namely, we adaptively generate a set of model parameters for a sequence of frames, and the sequence internally shares the same set of parameters to process each frame.

Specifically, as shown in Fig \ref{img_ade}, for the input sequence $\mathcal{V}=\{F_i \in \mathbb{R}^{3 \times H \times W}|i=1,2,…l\}$, we extract sequence features to generate parameter weights (expert weights) $w_i$. To reduce computational complexity, we perform equidistant sampling of $m=5$ frames instead of using all frames, with a sampling interval of $\lfloor l/m \rfloor$ ($\lfloor * \rfloor$ represents round down). Afterward, we extract degradation features from multiple sample frames as the degradation feature set $\mathcal{P}={p_i \in \mathbb{R}^{1 \times C}|i=1,2,…m\}}$ utilizing convolution, global average pooling, and linear mapping. $C$ represents the number of channels. We then concatenate all the feature vectors to obtain $P \in \mathbb{R}^{C\times m\times 1}$. We utilize learnable queries $Q \in \mathbb{R}^{C\times n\times 1}$  and cross-attention\cite{vaswani2017attention} to explore the information from each feature vector and combine them into a fixed-size feature vector. We then generate expert weights $W\in \mathbb{R}^{n\times 1}$ through a sigmoid function. The aforementioned process can be formulated as follows:
\begin{equation}
\begin{aligned}
W=Sigmoid\{L[Sofamax(\frac{L(Q)L(P)^T}{\sigma})L(P)]\},\label{eq_getweight}
\end{aligned}
\end{equation}
where $L(*)$ represents the linear layer. With this expert weight, we combine $n$ experts (learnable parameter $\theta_i$) to obtain $\Theta$:
\begin{equation}
\begin{aligned}
\Theta=\sum_{i=0}^nw_i\theta_i,\label{eq_expert}
\end{aligned}
\end{equation}
where $w_i$ is a component of $W$. Since this work involves All-In-One VR experiments for $4$ types of degradations, using four sets of expert parameters (i.e., $n=4$) is sufficient. The obtained $\Theta$ is used as the parameter for the U-Net, and the degradation matrices ($\mathcal{T}=\{T_i\in \mathbb{R}^{1 \times H \times W}|i=1,2,...,l\}$ and $\mathcal{D}=\{D_i\in \mathbb{R}^{1 \times H \times W}|i=1,2,...,l\}$) for the frame sequence $\mathcal{V}=\{F_i \in \mathbb{R}^{3 \times H \times W}|i=1,2,…l\}$ can be estimated:
\begin{equation}
\begin{aligned}
T_i,D_i=split[UNet(F_i;\Theta)],\label{eq_gettds}
\end{aligned}
\end{equation}
where $split$ indicates the operation of splitting along the channel dimension. The function of SADE can be summarized as
\begin{equation}
\begin{aligned}
\mathcal{T},\mathcal{D}=SADE(\mathcal{V}).\label{eq_sade}
\end{aligned}
\end{equation}
The SADE follows an end-to-end supervised training approach, and its loss function $\mathcal{L}_{SADE}$ is defined as:
\begin{equation}
\begin{aligned}
\mathcal{L}_{SADE}(\mathcal{T},\mathcal{D},\mathcal{V},\mathcal{Y})=\sum_{i=1}^l\mathcal{L}_{sup}[(F_i-D_i)/T_i,Y_i],\label{eq_adeloss}
\end{aligned}
\end{equation}
where $\mathcal{Y}=\{Y_i \in \mathbb{R}^{3 \times H \times W}|i=1,2,…,l\}$ represents the ground-truth video, and $L_{sup}$ has been defined in Eq. (\ref{eq_losssup}).

\subsection{Robust Warping Module}
\label{3.6}
\begin{figure}
	\begin{center}
		\includegraphics[width=\linewidth]{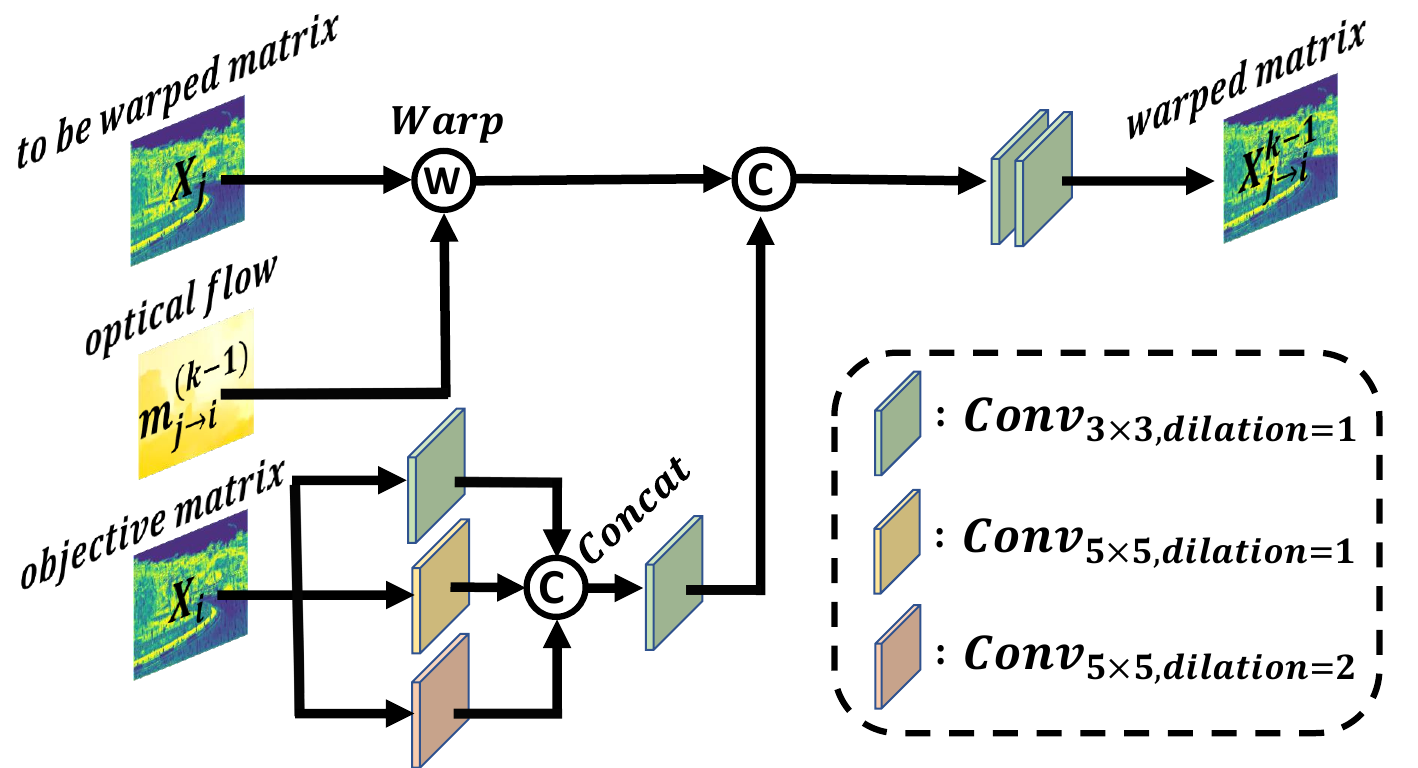}
	\end{center}
	\vspace{-15pt}
	\caption{Structure of Robust Warping Module (RWM). RWM implements explicit pixel motion based on optical flow and exhibits low motion errors.}
	\label{img_rwm}
\end{figure}
Frame alignment is crucial for utilizing information from adjacent frames. Existing explicit alignment methods\cite{shi2023videoflow,pan2020cascaded, wang2019edvr} are susceptible to large displacements or motion blur, which can negatively impact their precision. Although some implicit methods can mitigate this shortcoming \cite{li2023simple,sun2023deep}, this work relies on the explicit $M_{i\rightarrow j}$ operator and cannot utilize implicit methods. To address this issue, we propose a Robust Warping Module (RWM) to implement the $M_{i\rightarrow j}$ operator. The RWM retains the form of explicit alignment while overcoming the precision limitations of previous explicit alignment methods.

\begin{figure}
	\begin{center}
		\includegraphics[width=\linewidth]{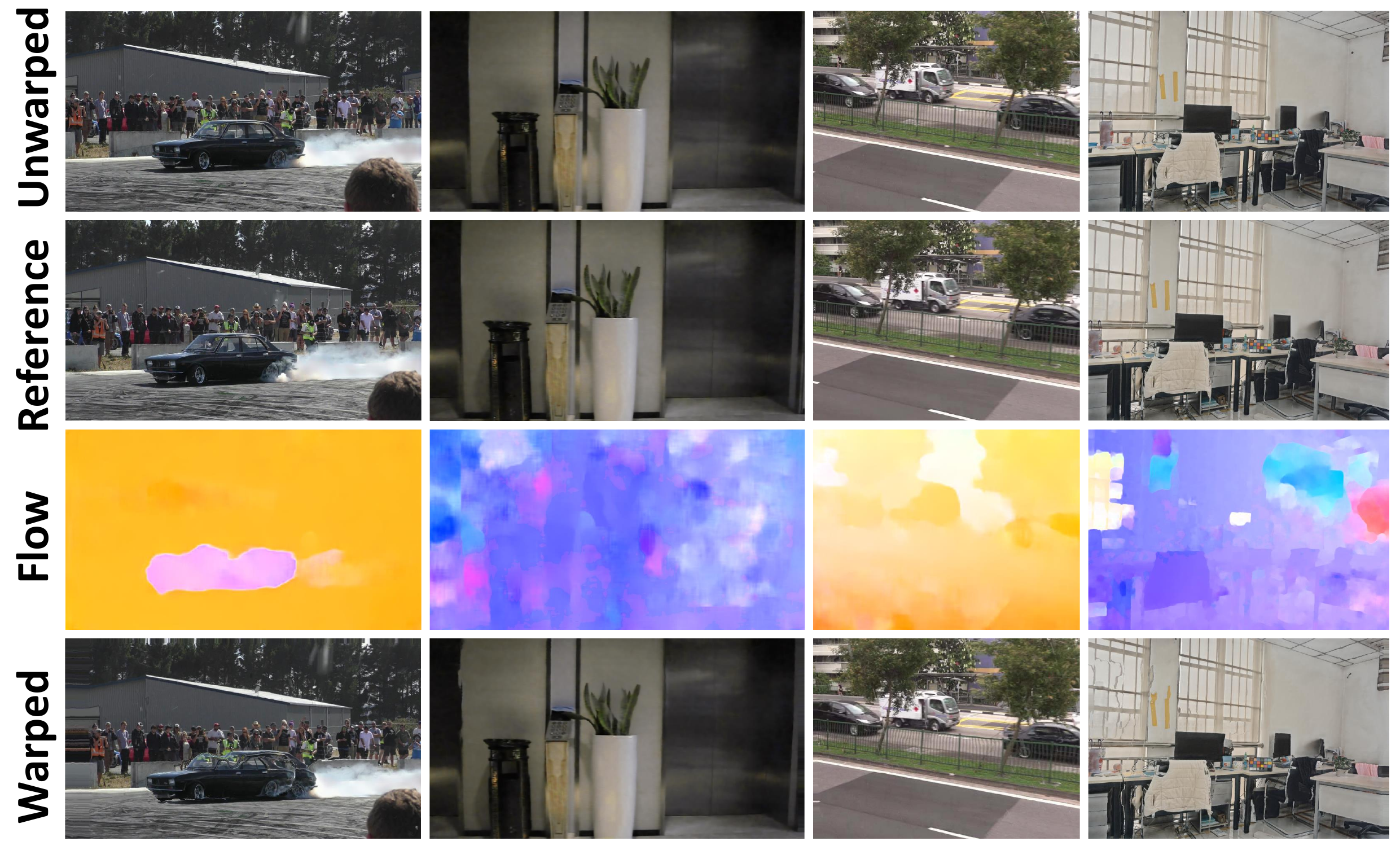}
	\end{center}
	\vspace{-15pt}
	\caption{Explicit optical flow-based warping operation may lead to structural errors. As shown in the figure, we applied warping operations to images using the optical flow estimated by FastFlowNet\cite{kong2021fastflownet}. It can be observed that the structural distortion is evident in the warped image. The proposed RWM alleviates this issue by incorporating the structural information of the objective image during warping operations.}
	\label{img_flows}
\end{figure}

The detailed structure of the Robust Warping Module (RWM) is illustrated in Fig. \ref{img_rwm}. Specifically, we correct potential structural errors (as shown in Fig. \ref{img_flows}) in the warped matrix by incorporating structural information from the objective matrix. From Eq. (\ref{eq_updatez}), it is clear that $M_{j\rightarrow i}$ directly operates on matrices $T_j(F_j-D_j)$ and $T_j^2$. Taking the former as an example, we first apply the $Warp_{j\rightarrow i}$ operation to matrix $T_j(F_j-D_j)$ based on the optical flow $m_{j\rightarrow i}$. According to Eq. (\ref{eq_allinone}), the spatial structure of $M_{j\rightarrow i}[T_j(F_j-D_j)]$ should be similar to $T_i(F_i-D_i)$ (objective matrix). Therefore, we propose to utilize $T_i(F_i-D_i)$ to correct structural errors in $ Warp _{j\rightarrow i}[T_j(F_j-D_j)]$. We extract the structural information from $T_i(F_i-D_i)$ using three convolutional layers ($3\times3$ with $dilation=1$, $5\times 5$ with $dilation=1$ and $5\times 5$ with $dilation=2$) with different receptive field sizes. Subsequently, we concatenate this structural information with $ Warp _{j\rightarrow i}[T_j(F_j-D_j)]$. And finally, two additional convolutional layers is used to fuse the information. The execution process of RWM can be formulated as follows (corresponding to Fig. \ref{img_rwm}):
%\begin{equation}
%\begin{aligned}
%RWM(x_j,x_i,m_{j\rightarrow i})=Warp(x_j,m_{j\rightarrow i}),\label{eq_rwm}
%\end{aligned}
%\end{equation}
\begin{subequations}
	\begin{align}
	RWM(X_j,X_i,m_{j\rightarrow i})=&Conv_{\times 2}\{Cat[Warp(X_j,\nonumber\\
	&m_{j\rightarrow i}),S(X_i)]\},\label{eq_rwm1}\\
	S(X_i) =& Conv\{Cat[Conv_{3\times 3}(X_i),\nonumber\\
	&Conv_{5\times 5,dilation=1}(X_i),\nonumber\\
	&Conv_{5\times 5,dilation=2}(X_i)]\},\label{eq_rwm2}
	\end{align}
\end{subequations}
where $X_j$ represents the matrix to be warped, $X_i$ denotes the objective matrix. $Conv$ denotes the convolutional layer. The arguments $size$ and $dilation$ for convolution defaults to $size=3\times 3$ and $dilation=1$ when there is no special annotation. $Cat$ signifies concatenating. $Warp$ denotes the process of displacing individual pixels based on the optical flow. Correspondingly, $M^{(k-1)}_{j\rightarrow i}[T_j(F_j-D_j)]$ and $M^{(k-1)}_{j\rightarrow i}(T_j^2)$ in Algorithm \ref{al_cdun} can be represented as:
\begin{subequations}
	\begin{align}
	M^{(k-1)}_{j\rightarrow i}[T_j(F_j-D_j)]=&RWM[T_j(F_j-D_j),\nonumber\\
	&T_i(F_i-D_i),m_{j\rightarrow i}^{(k-1)}],\label{eq_Ma}\\
	M^{(k-1)}_{j\rightarrow i}(T_j^2)=&RWM(T_j^2,T_i^2,m_{j\rightarrow i}^{(k-1)}).\label{eq_Mb}
	\end{align}
\end{subequations}
Note that we omit the objective matrix in Algorithm \ref{al_cdun} for representational convenience, without affecting its practical significance.
\begin{figure*}[t]
	\begin{center}
		\includegraphics[width=\linewidth]{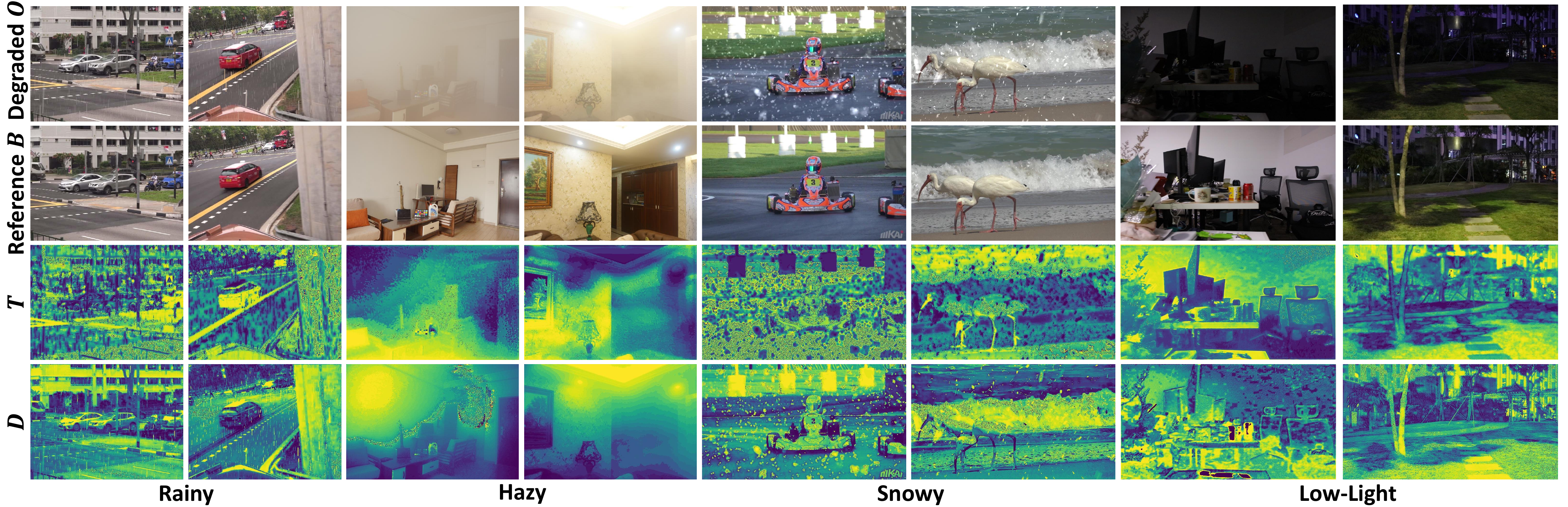}
	\end{center}
	\vspace{-15pt}
	\caption{The visualization displays the degradation matrices. The proposed SADE adaptively estimate the degradation matrices $T$ and $D$, which model the degradation type and features. It is worth noting that although the general degradation process (i.e., Eq. (\ref{eq_allinone})) includes a multiplicative factor and an additive factor, it merely aims to encompass various forms of degradation modeling. In practice, $T$ lacks practical significance in rain modeling, while $D$ lacks practical significance in low-light modeling.}
	\label{img_tds}
\end{figure*}
\subsection{Spatial Motion Weight}
\label{3.7}
Generally, long-distance motion often exhibit larger errors compared to short-distance displacements in the process of frame alignment. Hence, pixels that have moved different distances in adjacent frames contribute differently to the restoration of the current frame. In order to mitigate the impact of motion errors on the fusion of adjacent frame information, we allocate weights (i.e., $W_{i\rightarrow j}$ in Eq. (\ref{eq_energyv})) to each pixel based on the distance it has moved. Since the optical flow contains the coordinates of the pixel movement, this movement distance can be calculated from the optical flow.

Specifically, we propose to use a learnable network to estimate spatial motion weight $W_{i\rightarrow j}$ based on optical flow $m_{i\rightarrow j}$. Moreover, for a simpler calculation, we directly calculate $U_{i\rightarrow j}$ instead of $W_{i\rightarrow j}$ to reduce the frequency of calls to the above-mentioned learnable network and motion operator $M_{j\rightarrow i}$ (see supplementary material for the detailed derivation and analysis):
\begin{equation}
\begin{aligned}
U_{i\rightarrow j} =\mathcal{F}(||m_{i\rightarrow j}||^2),\label{eq_getU}
\end{aligned}
\end{equation}
where $\mathcal{F}$ is a learnable network with a $convolution\rightarrow normalization\rightarrow ReLU\rightarrow convolution\rightarrow Sigmoid$ structure.

\section{Experiments and Analysis}

In this section, we conducted experimental evaluation of the proposed method. Firstly, we described the experimental settings, followed by comparisons with existing state-of-the-art methods on both All-In-One VR and Task-Specific VR. Finally, we performed relevant ablation studies to demonstrate the effectiveness of the proposed key modules.
\label{4}
\begin{figure*}
	\begin{center}
		\includegraphics[width=\linewidth]{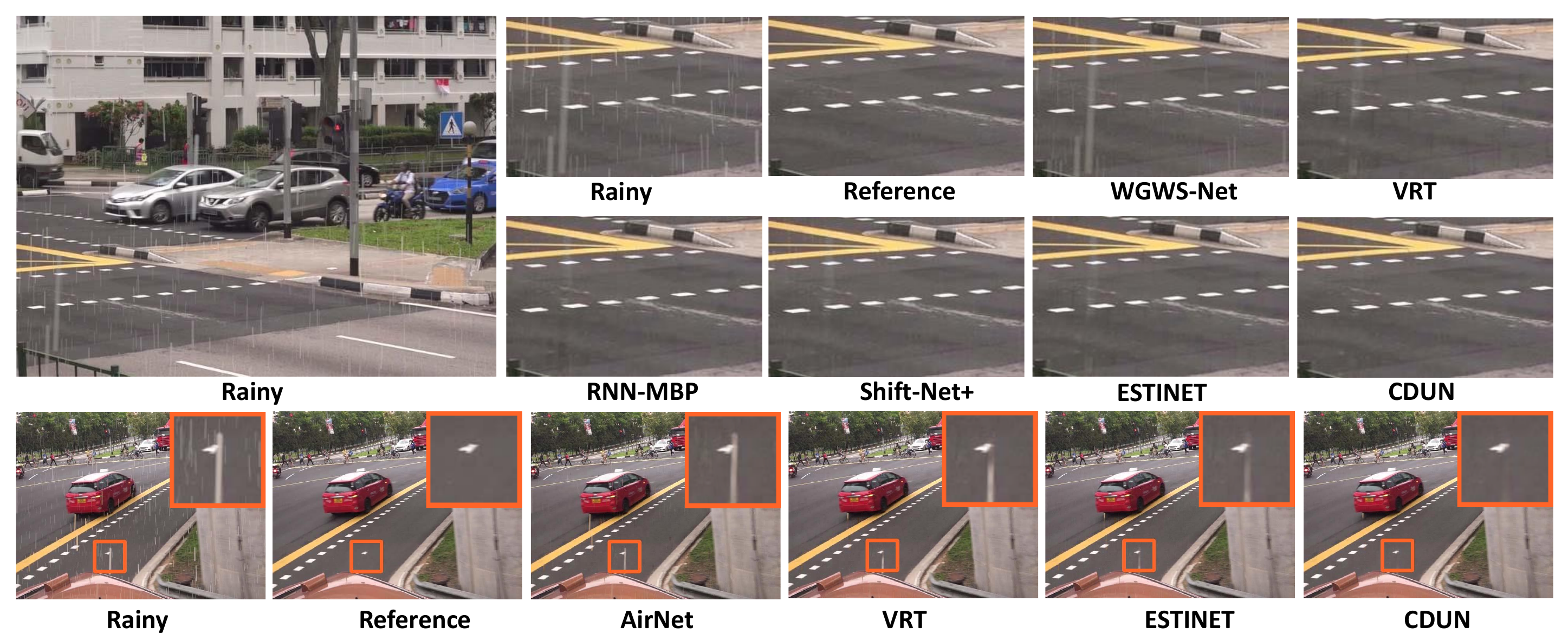}
	\end{center}
	\vspace{-15pt}
	\caption{Visual comparison of video rain removal results on the NTURain dataset\cite{chen2018robust} among different methods. The proposed CDUN restores a cleaner background frame compared to other methods.}
	\label{img_rainrm}
\end{figure*}
\subsection{Experiment settings}
\label{4.1}
\textbf{Datasets}. We evaluate the VR performance on four tasks: rain removal, haze removal, snow removal, and low-light enhancement (LLE). The detailed information of the training and testing datasets is presented in Tab. \ref{tab_dataset}. Due to the lack of publicly available paired datasets of snow videos and clean videos, we utilized the DAVIS (2017 challenge, 480p) video dataset\cite{pont20172017} and Adobe After Effects to synthesize snow videos and construct a paired dataset. The training set used in the All-In-One VR experiments is a combination of the four individual training sets.
\begin{table}[h]
	\vspace{-2mm}
	\begin{center}
		\caption{The detailed information about the datasets used in this work.}\label{tab_dataset}
		\renewcommand\arraystretch{1.2}
		\tabcolsep=0.04cm
				\vspace{-5pt}
		\begin{tabular}{@{}|c|c|c|c|@{}}
			\hline
			\multirow{2}*{Task}&\multirow{2}*{Dataset}&Number of Training&Number of Testing\\
			&&Videos / Frames& Videos / Frames\\
			\hline
			Rain Removal&NTURain\cite{chen2018robust}&$25 / 3.1k$&$8 / 1.7k$\\
			Haze Removal&REVIDE\cite{zhang2021learning}&$42 / 1.7k$&$6 / 0.3k$\\
			Snow Removal&DAVIS-snow\cite{pont20172017}&$90 / 6.2k$&$30 / 2.1k$\\
			LLE&SDSD\cite{wang2021seeing}&$125 / 32.1k$&$25 / 5.4k$\\
			\hline
		\end{tabular}
	\end{center}
\end{table}

\begin{table*}[t]
	\begin{center}
		\caption{Quantitative comparison with state-of-the-art methods on All-In-One VR. In the table, the best score is emphasized in \textbf{bold}.}
		\label{tab_allinone}
		\tabcolsep=0.10cm
		\renewcommand\arraystretch{1.4}
		\vspace{-5pt}
		\begin{tabular}{@{}c|c|cccc|cccccc|c@{}}
			\hline
			\hline
			&\multirow{2}*{Type}&\multicolumn{4}{c}{\multirow{2}*{All-In-One IR Method}}&\multicolumn{6}{c}{\multirow{2}*{Task-Specific VR Method}}&\multirow{2}*{\shortstack{All-In-One VR\\Method}}\\
			&&&&&&&&&&&&\\
			%\multirow{2}*{Metrics}
			\cline{2-13}
			&\multirow{2}*{Method}&TransWeather&Chen et al.&AirNet&WGWS-Net&STFAN&SDSD&VRT&RNN-MBP&ESTINET&Shift-Net+&CDUN\\
			&&\cite{valanarasu2022transweather}&\cite{chen2022learning}&\cite{li2022all}&\cite{zhu2023learning}&\cite{zhou2019spatio}&\cite{wang2021seeing}&\cite{liang2022vrt}&\cite{zhu2022deep}&\cite{zhang2022enhanced}&\cite{li2023simple}&(Ours)\\
			\hline
			\multirow{2}*{\shortstack{Rain\\Removal}}&PSNR $\uparrow$&30.16&31.79&31.82&32.07&31.25&32.79&34.25&34.32&35.49&34.99&\textbf{37.16} (+1.67)\\
			&SSIM $\uparrow$&0.923&0.936&0.941&0.945&0.936&0.946&0.952&0.951&0.956&0.953&\textbf{0.969} (+0.013)\\
			\hline
			\multirow{2}*{\shortstack{Hazy\\Removal}}&PSNR $\uparrow$&18.27&21.12&21.21&21.93&19.76&21.99&22.34&22.21&22.52&22.63&\textbf{23.54} (+0.91)\\
			&SSIM $\uparrow$&0.837&0.855&0.860&0.868&0.843&0.867&0.886&0.881&0.888&0.890&\textbf{0.899} (+0.009)\\
			\hline
			\multirow{2}*{\shortstack{Snow\\Removal}}&PSNR $\uparrow$&28.96&29.73&30.32&30.57&29.24&29.72&31.17&31.30&31.46&31.82&\textbf{33.78} (+1.96)\\
			&SSIM $\uparrow$&0.881&0.895&0.905&0.911&0.886&0.892&0.913&0.912&0.917&0.925&\textbf{0.942} (+0.017)\\
			\hline
			\multirow{2}*{LLE}&PSNR $\uparrow$&20.17&21.61&22.03&22.05&21.10&23.12&23.18&23.26&23.28&23.31&\textbf{24.18} (+0.87)\\
			&SSIM $\uparrow$&0.628&0.633&0.636&0.642&0.625&0.654&0.655&0.660&0.663&0.676&\textbf{0.694} (+0.018)\\
			\hline
			\multirow{2}*{Average}&PSNR $\uparrow$&24.39&26.06&26.34&26.65&25.34&26.91&27.73&27.77&28.19&28.19&\textbf{29.67} (+1.48)\\
			&SSIM $\uparrow$&0.817&0.830&0.836&0.842&0.823&0.840&0.852&0.851&0.856&0.861&\textbf{0.876} (+0.015)\\
			\hline
			\hline
		\end{tabular}
	\end{center}
\end{table*}
\begin{figure*}
	\begin{center}
		\includegraphics[width=\linewidth]{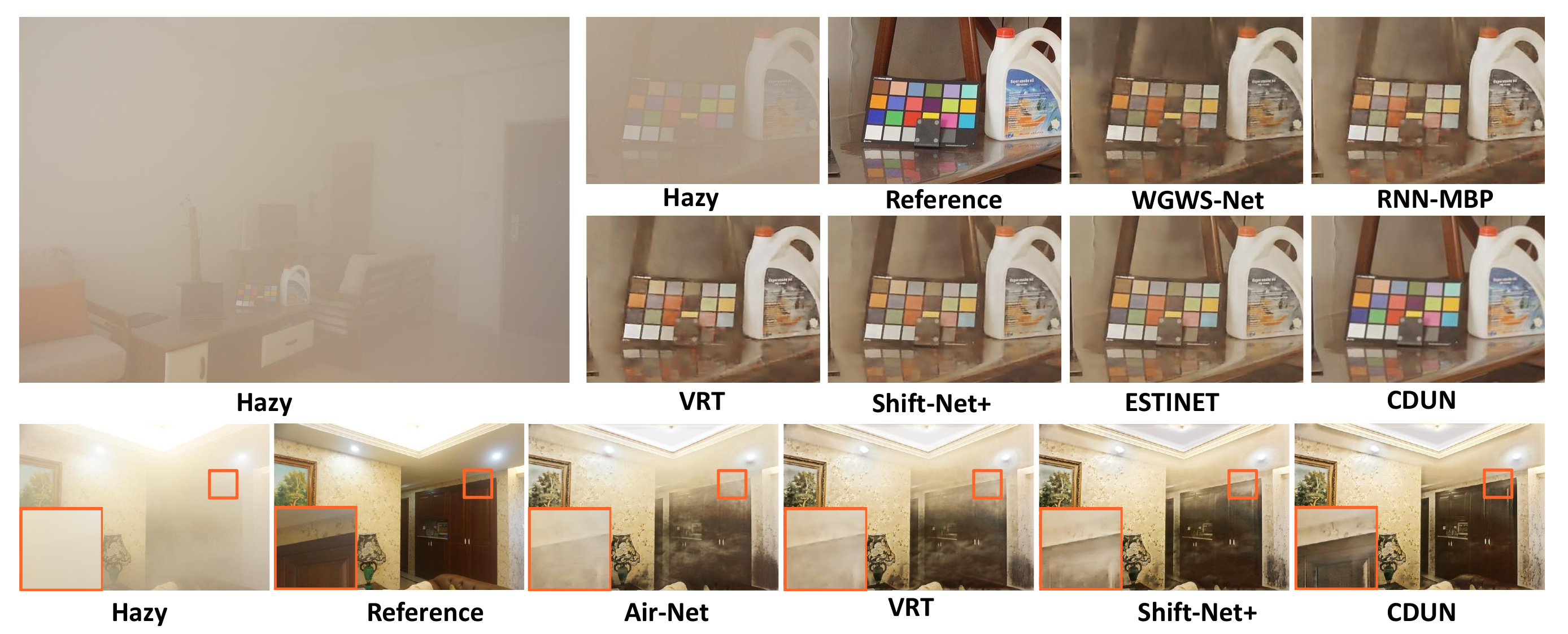}
	\end{center}
	\vspace{-15pt}
	\caption{Visual comparison of video haze removal results on the REVIDE dataset \cite{zhang2021learning} among different methods. The proposed CDUN can captures degradation features such as haze density, resulting in fewer artifacts in the restored output.}
	\label{img_hazerm}
\end{figure*}

\textbf{Training}. The training process in this work includes two stages: training of SADE and training of CDUN. Both stages utilize the same training datasets. During the data sampling process from the training dataset, we randomly select a contiguous sequence of frames with a length of $l$ from the four degradation types with equal probability. $l$ is set to $7$ during the training of SADE and to $3$ during the formal training of the overall model. The spatial scale of the training frame sequences is randomly cropped to a size of $256\times 256$, and data augmentation is performed by applying random horizontal flipping. In both training stages, we utilize the AdamW optimizer with a batch size of $4$ and an initial learning rate of $2e-4$. The learning rate is reduced by a factor of one-tenth every $40k$ iterations, and a total of $100k$ iterations are performed.

\textbf{Evaluation}. 
We employ SSIM (Structural Similarity Index) \cite{wang1987ssim} and PSNR (Peak Signal-to-Noise Ratio) as evaluation metrics to assess the video restoration quality. Additionally, we evaluate the computational cost of video restoration methods based on FLOPs (Floating-Point Operations), parameter quantity, and runtime. The FLOPs and runtime presented in the tables of this paper represent the average computation and runtime per frame. $\uparrow$ in the table indicates that the higher the score, the better the performance.

\subsection{Comparisons on All-In-One Video Restoration}
\label{4.2}

\begin{figure*}
	\begin{center}
		\includegraphics[width=\linewidth]{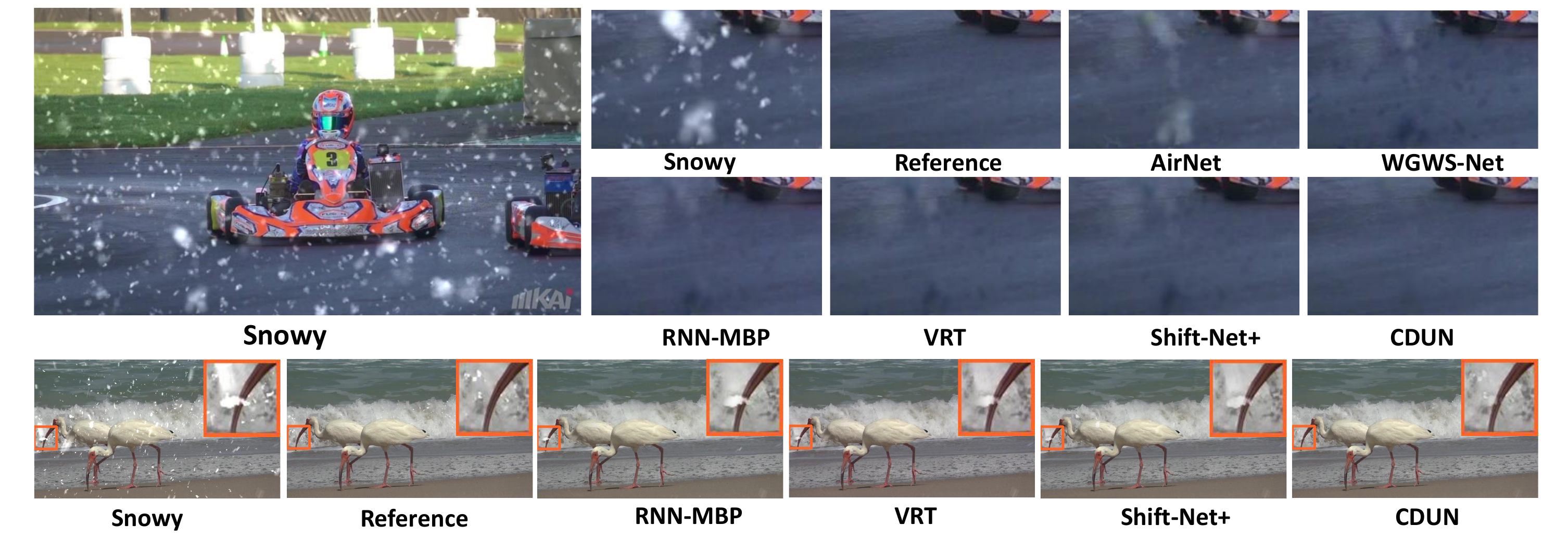}
	\end{center}
	\vspace{-15pt}
	\caption{Visual comparison of video snow removal results on the DAVIS-snow dataset \cite{pont20172017} among different methods. It can be seen from the figure that the proposed CDUN achieves outstanding results in removing large snowflakes.}
	\label{img_snowrm}
\end{figure*}

In this work, we compared the proposed CDUN with existing state-of-the-art methods on the All-In-One VR task. Tab \ref{tab_allinone} presents the scores of different methods on multiple datasets. As our work is the first investigation into All-In-One VR, we compared it with some excellent existing All-In-One IR and Task-Specific VR methods. As shown in Tab \ref{tab_allinone}, the proposed method outperforms other methods on all four tasks. Specifically, compared to the second-best method, the proposed approach achieves an average PSNR improvement of 1.48dB and an SSIM improvement of 0.015. %Furthermore, as shown in Fig. \ref{img_param}, the proposed CDUN exhibits a significant advantage in terms of parameter quantity compared to existing methods.

\begin{figure*}
	\begin{center}
		\includegraphics[width=\linewidth]{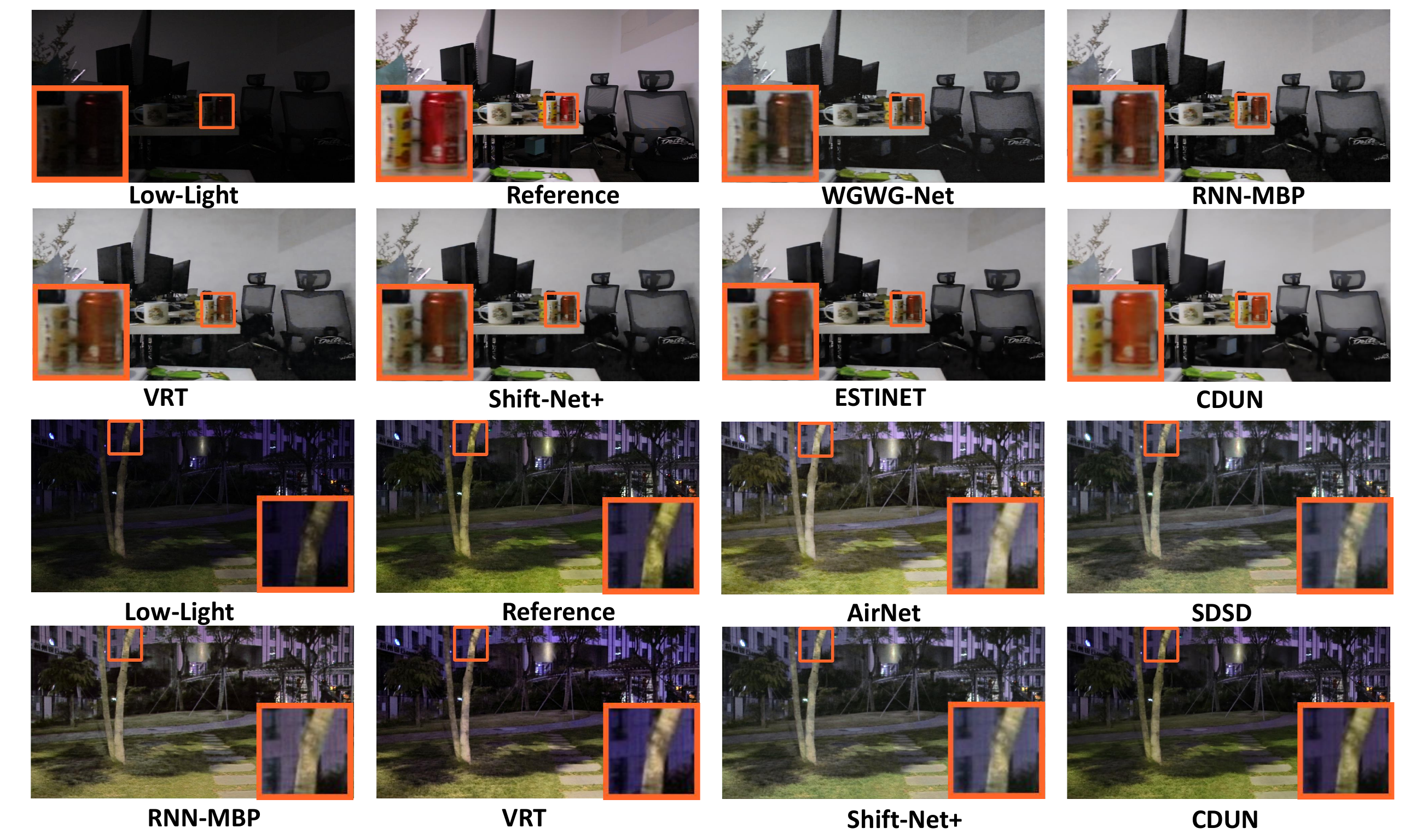}
	\end{center}
	\caption{Visual comparison of video low-light enhancement results on the SDSD dataset \cite{wang2021seeing} among different methods. The proposed CDUN achieves more visually pleasing results compared to existing methods.}
	\label{img_lle}
\end{figure*}

Furthermore, illustrated in Fig. \ref{img_rainrm}, \ref{img_hazerm}, \ref{img_snowrm}, and \ref{img_lle}, we present qualitative comparisons of the proposed CDUN with existing methods on video rain removal, haze removal, snow removal, and low-light enhancement tasks. It is evident that, the videos restored by CDUN exhibit more visually pleasing results compared to other methods.

The proposed CDUN has been evaluated through quantitative and qualitative comparisons, demonstrating its significant advantages over existing methods in All-In-One VR. Additionally, we visualized the degradation matrices estimated by SADE in the All-In-One VR task, and they are presented in Fig. \ref{img_tds}. Indeed, the degradation matrices model different degradation features, such as rain streaks, haze density, snowflake, and lighting intensity. The proposed CDUN achieves adaptive removal of different degradations through guidance from different degradation matrices.

\subsection{Comparisons on Task-Specific Video Restoration}
\label{4.3}
\begin{table*}[t]
	\begin{center}
		\caption{Quantitative comparison with state-of-the-art methods on Task-Specific VR. In the table, the best score is emphasized in \textbf{bold}.}
		\label{tab_taskspecific}
		\tabcolsep=0.03cm
		\renewcommand\arraystretch{1.4}
		\vspace{-5pt}
		\begin{tabular}{@{}c|c|cccccccc|c@{}}
			\hline
			\hline
			\multirow{3}*{\shortstack{Rain\\Removal}}&Method&MS-CSC\cite{li2018video}&JORDER\cite{yang2017deep}&FastDerain\cite{jiang2018fastderain}&J4RNet\cite{liu2018erase}&SPAC\cite{chen2018robust}&FCRNet\cite{yang2019frame}&SLDNet+\cite{yang2022learning}&ESTINET\cite{zhang2022enhanced}&CDUN\\
			&PSNR$\uparrow$ &27.31&32.61&30.32&32.14&33.11&36.05&38.36&37.48&\textbf{38.39}\\
			&SSIM$\uparrow$ &0.7870&0.9482&0.9262&0.9480&0.9474&0.9676&0.9750&0.9700&\textbf{0.9760}\\
			\hline
			\multirow{3}*{\shortstack{Haze\\Removal}}&Method&MSBDN\cite{dong2020multi}&FFA\cite{luo2021global}&VDH\cite{ren2018deep}&EDVR\cite{wang2019edvr}&CG-IDN\cite{zhang2021learning}&NCFL\cite{huang2022neural}&BasicVSR++\cite{chan2022basicvsr++}&MAP-Net\cite{xu2023video}&CDUN\\
			&PSNR $\uparrow$&22.01&16.65&16.64&21.22&23.21&23.63&21.68&24.16&\textbf{24.31}\\
			&SSIM $\uparrow$&0.8759&0.8133&0.8133&0.8707&0.8836&0.8925&0.8726&0.9043&\textbf{0.9109}\\
			\hline
			\multirow{3}*{\shortstack{Snow\\Removal}}&Method&MS-CSC\cite{li2018video}&SPAC\cite{chen2018robust}&FCR-Net\cite{yang2019frame}&MAP-Net\cite{xu2023video}&SLDNet+\cite{yang2022learning}&RNN-MBP\cite{zhu2022deep}&ESTINET\cite{zhang2022enhanced}&Shift-Net+\cite{li2023simple}&CDUN\\
			&PSNR $\uparrow$&29.07&29.90&30.39&31.77&32.60&33.24&33.14&33.85&\textbf{34.73}\\
			&SSIM $\uparrow$&0.8874&0.9025&0.9081&0.9202&0.9293&0.9371&0.9408&0.9447&\textbf{0.9613}\\
			\hline
			\multirow{3}*{LLE}&Method&DeepUPE\cite{wang2019underexposed}&DeepLPF\cite{moran2020deeplpf}&DRBN\cite{yang2020fidelity}&MBLLEN\cite{lv2018mbllen}&SMID\cite{chen2019seeing}&SMOID\cite{jiang2019learning}&StableLLVE\cite{zhang2021learning}&SDSD\cite{wang2021seeing}&CDUN\\
			&PSNR $\uparrow$&21.91&22.14&22.23&21.48&23.74&23.17&23.41&24.23&\textbf{24.58}\\
			&SSIM $\uparrow$&0.66&0.63&0.63&0.61&0.69&0.65&0.67&\textbf{0.70}&\textbf{0.70}\\
			\hline
			\hline
		\end{tabular}
	\end{center}
\end{table*}

Since there are currently only task-Specific VR methods and this work is the first work about All-In-One VR, we also conducted Task-Specific VR experiments for direct comparisons with existing methods. Following the works of Zhang et al.\cite{zhang2022enhanced}, Xu et al.\cite{xu2023video}, Yang et al.\cite{yang2022learning} and Wang et al.\cite{wang2021seeing}, we trained the proposed method using the same experimental settings as theirs. Subsequently, we conducted quantitative comparisons on video rain removal, haze removal, snow removal, and low-light enhancement tasks. As shown in Tab. \ref{tab_taskspecific}, the proposed CDUN achieved the highest performance scores in all tasks. This is because the iterative process of CDUN expands the temporal receptive fields, enabling the fusion of more information from additional frames to help with the restoration of corrupted content. This mechanism is beneficial for both All-In-One and Task-Specific VR tasks.
\subsection{Ablation Study}
\label{4.4}
In this section, we conducted ablation studies about key modules to explore their contributions and functionalities. All experiments are conducted on the All-In-One VR tasks. The scores of the best-performing results are highlighted in \textbf{bold} in the tables.

\textbf{The effect of SADE.} Our proposed CDUN utilizes SADE to adaptively estimate the degradation matrices, thereby guiding the optimization process (i.e. Eq. (\ref{eq_energyv})) to adaptively remove the corresponding type of degradation. SADE achieves the adaptive adjust of the network parameters based on the degradation characteristics present in the input video, enhancing the accuracy of degradation modeling in multi-degradation scenarios. Furthermore, the proposed SADE utilizes a sequence-wise adaptive mechanism instead of the classical frame-wise scheme, effectively overcoming the instability issues associated with parameter generation in the latter. Due to the existing Task-Specific IR methods has directly used UFormer\cite{wang2022uformer} to predict the degradation matrices\cite{guo2023shadowdiffusion}, we conducted comparative experiments by replacing SADE with UFormer. As shown in the first row and the last row of Tab. \ref{tab_abthree}, employing SADE leads to a significant improvement in PSNR. This demonstrates that, compared to the proposed SADE, the static model (UFormer) is limited in dealing with multi-degradation scenarios effectively.

\begin{table}[h]
	\begin{center}
		\caption{The ablation study on SADE, RWM, and spatial motion weight. The table presents the performance scores (PSNR) of All-In-One VR under various experimental configurations.}
		\label{tab_abthree}
		\tabcolsep=0.04cm
		\renewcommand\arraystretch{1.4}
		\vspace{-5pt}
		\begin{tabular}{@{}cccccccc@{}}
			\hline
			\hline
			\multirow{2}*{SADE}&\multirow{2}*{RWM}&\multirow{2}*{\shortstack{Spatial\\Motion Weight}}&\multirow{2}*{\shortstack{Rain\\Removal}}&\multirow{2}*{\shortstack{Haze\\Removal}}&\multirow{2}*{\shortstack{Snow\\Removal}}&\multirow{2}*{LLE}&\multirow{2}*{Average}\\
			&&&&&&&\\
			\hline
			&\checkmark&\checkmark&35.27&22.64&32.92&22.30&28.28\\
			\checkmark&&\checkmark&32.26&20.07&27.96&20.76&25.26\\
			\checkmark&\checkmark&&36.85&23.08&31.99&22.66&28.65\\
			\checkmark&\checkmark&\checkmark&\textbf{37.46}&\textbf{23.54}&\textbf{33.78}&\textbf{24.18}&\textbf{29.74}\\
			\hline
			\hline
		\end{tabular}
	\end{center}
\end{table}

\begin{table}[h]
	\begin{center}
		\caption{Ablation study on Sequence-wise adaptive and frame-wise adaptive strategies. The table presents a comparison of VR performance under different experimental settings, based on PSNR scores.}
		\label{tab_sworfw}
		\tabcolsep=0.2cm
		\renewcommand\arraystretch{1.4}
		\vspace{-5pt}
		\begin{tabular}{@{}cccccc@{}}
			\hline
			\hline
			\multirow{2}*{Method}&\multirow{2}*{\shortstack{Rain\\Removal}}&\multirow{2}*{\shortstack{Haze\\Removal}}&\multirow{2}*{\shortstack{Snow\\Removal}}&\multirow{2}*{LLE}&\multirow{2}*{Average}\\
			&&&&&\\
			\hline
			frame-wise&34.51&21.61&32.09&23.06&27.82\\
			sequence-wise&\textbf{37.46}&\textbf{23.54}&\textbf{33.78}&\textbf{24.18}&\textbf{29.74}\\
			\hline
			\hline
		\end{tabular}
	\end{center}
\end{table}
\begin{figure}
	\begin{center}
		\includegraphics[width=\linewidth]{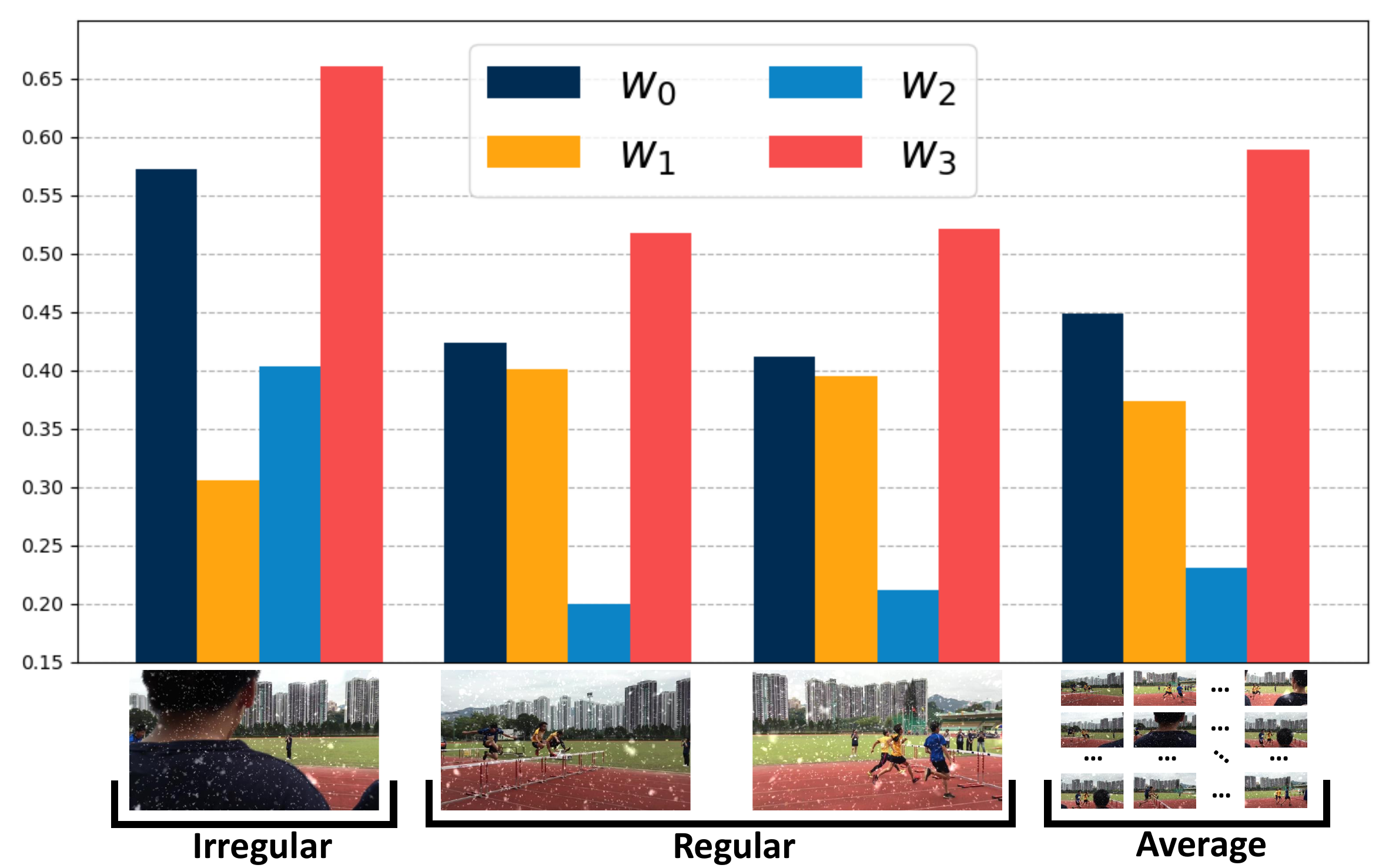}
	\end{center}
	\vspace{-15pt}
	\caption{The visualization of expert weights in SADE under the frame-wise strategy. The bar chart displays the values of the four expert weights. In the frame-wise setting, when the background information differs significantly from other frames (as seen in the leftmost column), there are considerable variations in expert weights. This background sensitivity hinders the adaptive estimation of the degradation matrices.}
	\label{img_dyw}
\end{figure}
\begin{table}[h]
	\begin{center}
		\caption{SADE generates adaptive parameters by observing information from $m$ frames. The table below showcases the PSNR scores and FLOPs for different values of $m$.}
		\label{tab_nframes}
		\renewcommand\arraystretch{1.4}
		\vspace{-5pt}
		\begin{tabular}{@{}ccccccc@{}}
			\hline
			\hline
			\multirow{2}*{$m$}&\multirow{2}*{\shortstack{Rain\\Removal}}&\multirow{2}*{\shortstack{Haze\\Removal}}&\multirow{2}*{\shortstack{Snow\\Removal}}&\multirow{2}*{LLE}&\multirow{2}*{Average}&\multirow{2}*{FLOPs (G)}\\
			&&&&&&\\
			\hline
			1&33.79&22.34&30.68&22.14&27.24&2.153\\
			2&34.45&22.46&31.90&23.07&27.97&2.448\\
			3&36.34&23.22&33.29&23.43&29.07&2.965\\
			5&\textbf{37.46}&\textbf{23.54}&\textbf{33.78}&\textbf{24.18}&\textbf{29.74}&3.953\\
			\hline
			\hline
		\end{tabular}
	\end{center}
\end{table}

\begin{figure}
	\begin{center}
		\includegraphics[width=\linewidth]{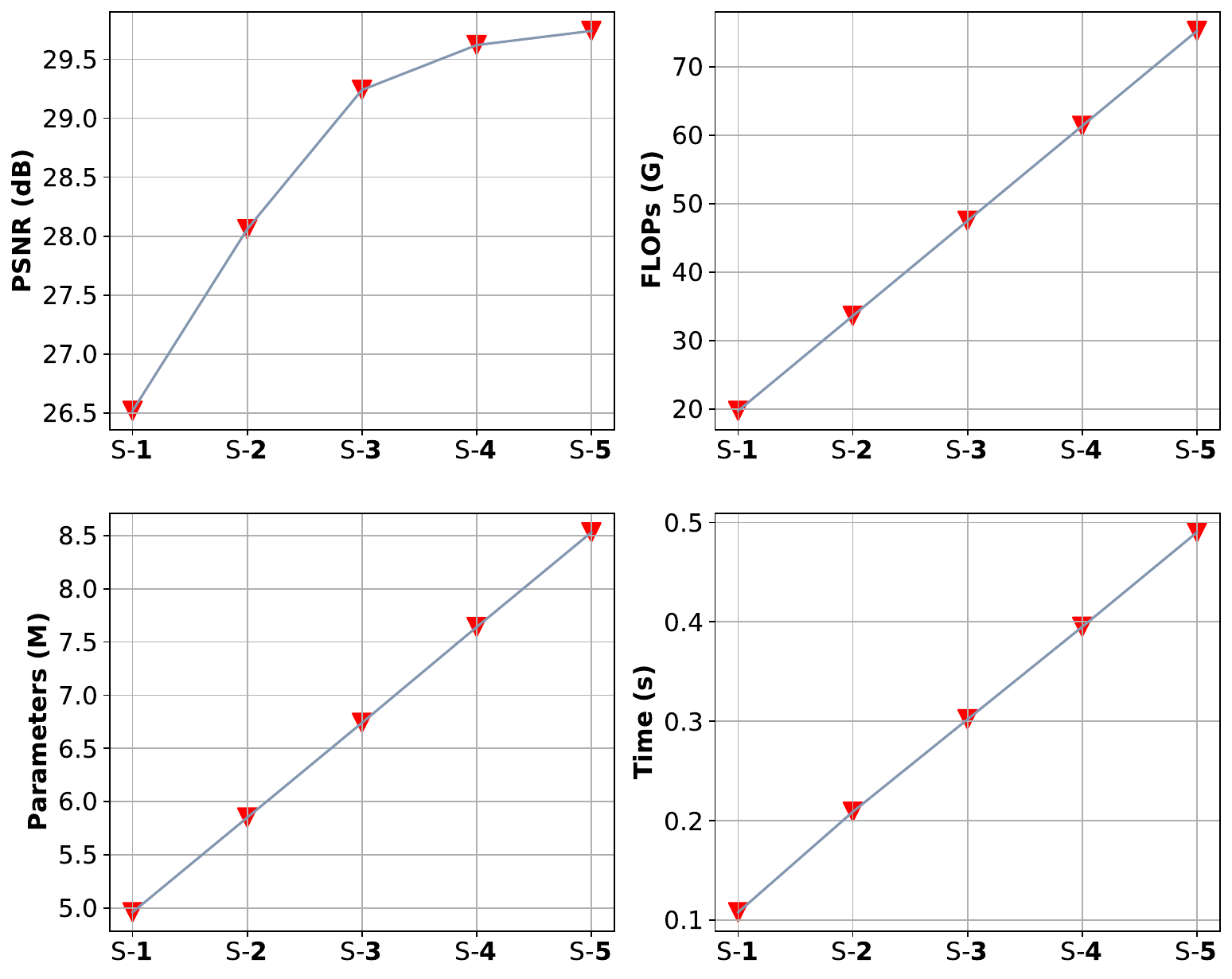}
	\end{center}
	\vspace{-15pt}
	\caption{The performance scores (PSNR and SSIM) and computational cost (FLOPs, parameter quantity, and runtime) are qualitatively compared in different numbers of iteration steps.}
	\label{img_lineitn}
\end{figure}

\begin{table}[h]
	\begin{center}
		\caption{The performance scores (PSNR and SSIM) and computational cost (FLOPs, parameter quantity, and runtime) are quantitatively compared in different numbers of iteration steps. The $avg$ represents the average score across the four tasks: rain removal, haze removal, snow removal, and low-light enhancement.}
		\label{tab_itn}
		\tabcolsep=0.12cm
		\renewcommand\arraystretch{1.4}
		\vspace{-5pt}
		\begin{tabular}{@{}cccccc@{}}
			\hline
			\hline
			$N$&PSNR($avg$)&SSIM($avg$)&FLOPs (G)&$\#$Params (M)&Runtime (s)\\
			\hline
			1&26.52&0.823&19.74&4.96&0.108\\
			2&28.06&0.847&33.61&5.85&0.209\\
			3&29.24&0.865&47.49&6.74&0.302\\
			4&29.62&0.871&61.36&7.64&0.395\\
			5&\textbf{29.74}&\textbf{0.876}&75.23&8.53&0.490\\
			\hline
			\hline
		\end{tabular}
	\end{center}
\vspace{-10pt}
\end{table}

\textbf{Advantages of Sequence-wise Adaptation over Frame-wise Adaptation.} This work introduces a SADE for adaptive estimating various types of degradation matrices. The conventional frame-wise approach involves generating network parameters based on the features of each frame to process that particular frame. However, this is not suitable for estimating degradation matrices in All-In-One VR. As shown in Fig. \ref{img_dyw}, we trained a degradation estimator using the frame-wise architecture and visualized the expert weights (denoted as $w_i$ in Eq. (\ref{eq_expert}) and Fig. \ref{img_ade}). To obtain a reference, we calculated the average of the weights from multiple frames, as shown in the rightmost column of Fig. \ref{img_dyw}. We observed that when the background scene of individual frames significantly differs from that of the majority of frames (the leftmost column of Fig. \ref{img_dyw}), the generated expert weights deviate significantly from the reference values as well. Our objective is to enable the degradation estimator to utilize different parameters for different degradation types, but the parameter adjustment in the frame-wise architecture is highly sensitive to background information. Therefore, the model tends to waste its representation capacity on processing the background, leading to a decrease in the estimation accuracy of the degradation matrices.

To address the aforementioned issues, we propose the sequence-wise adaptive strategy, which involves generating network parameters by observing multiple frames within a video clip and sharing these parameters while processing individual frames within that video. Typically, degradation within a video clip is of the same type. Therefore, the introduced sequence-wise adaptive strategy overcomes the limitations of the frame-wise approach without compromising the model's adaptive nature. Tab. \ref{tab_sworfw} presents a quantitative comparison between the two strategies. Additionally, we conducted an ablation study on the number of frames ($m$) used for parameter generation, and the relevant experimental results are provided in Tab. \ref{tab_nframes}. The final SADE model generates parameters by observing $5$ frames, uniformly sampled at equal intervals.

\textbf{The effect of RWM}. This work implements the $M_{i\rightarrow j}$ operator via RWM. RWM overcomes the limited accuracy drawback of traditional explicit alignment methods while preserving the essential explicit alignment paradigm. From Fig. \ref{img_flows}, it can be observed that due to the presence of errors in optical flow estimation, direct alignment based on optical flow may result in structural distortions. To address this issue, we introduced a corrective step after the explicit warping operation. This step involves extracting structural information from the objective matrix (see Sec. \ref{3.6} for details) to rectify the results, aiming to mitigate the adverse effects of optical flow inaccuracies and reduce structural distortions in the aligned output. The experimental results in row 2 and 4 of Tab. \ref{tab_abthree} reveal that the proposed RMW plays a crucial role.

\textbf{The effect of spatial motion weight}. Generally, during the process of video alignment, pixels that are displaced over longer distances tend to have larger optical flow estimation errors. As a result, pixels displaced with different distances have different contributing levels for frame restoration. In this work, we address this issue by assigning different weights to pixels with different displacement distances, thereby enhancing the model's tolerance toward the errors of optical flow estimating. As evident from the comparison between rows 2 and 4 in Tab. \ref{tab_abthree}, it can be observed that the proposed spatial motion weight is crucial for the restoration performance.

\textbf{Number of iterations}. The proposed CDUN performs All-In-One VR by iteratively optimizing Eq. (\ref{eq_energyv}) for a fixed number of steps. To determine the appropriate value of the iteration steps $N$, we conducted experiments to find the optimal value for $N$. As shown in Tab. \ref{tab_itn}, with an increase in the number of iteration steps, the PSNR, FLOPs, and parameter quantity of CDUN gradually increase. From Fig. \ref{img_lineitn}, it can be observed that the acceleration of PSNR improvement decreases progressively, while the growth in computational cost follows a linear trend with a almost constant acceleration. This reflects that in later stages, further increasing the number of iteration steps has a lower cost-effectiveness in terms of performance improvement. Therefore, to trade-off the computational cost and model performance, we ultimately set the number of iteration steps to $5$. Additionally, the temporal window length for each iteration step is set to $3$. Consequently, for five iteration steps, the total temporal receptive field length can be calculated as $3 + (5-1) \times (3-1) = 11$. Such a temporal receptive field length is relatively sufficient for video restoration.

\section{Disscussion}
\begin{figure}
	\begin{center}
		\includegraphics[width=\linewidth]{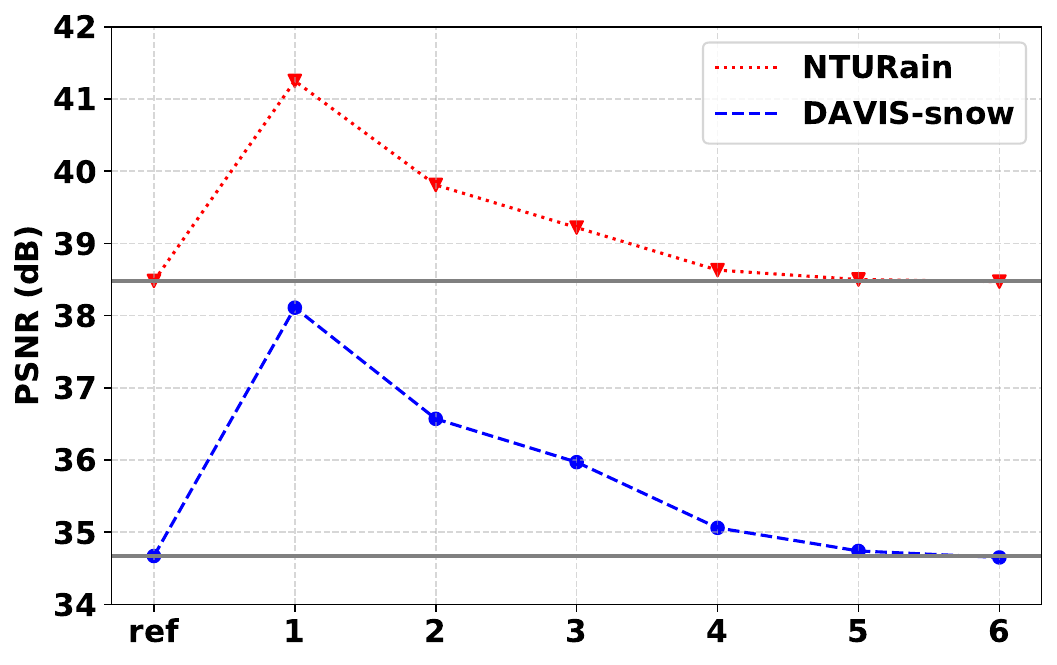}
	\end{center}
	\vspace{-15pt}
	\caption{In this experiment, we explore the temporal receptive field length by replacing adjacent frames with ground-truth frames and observing the performance score improvements. Here, $ref$ represents the PSNR score without inserting ground-truth frames. The remaining horizontal coordinates represent the insertion position, denoted by the value of $j$.}
	\label{img_recfield}
\end{figure}
\textbf{Study on the Expansion of Temporal Receptive Field and Its Impact.} In this experiment, we explore the capability of the proposed CDUN in expanding the temporal receptive field by stacking temporal windows. The experiment is conducted using the middle $13$ frames of each video clip from the NTURain \cite{chen2018robust} dataset and DAVIS-snow\cite{pont20172017} dataset. We number the middle frame as $i$ and subsequently labeled the $13$ frames as $i-6$, $i-5$, ..., $i$, ..., $i+5$, $i+6$. Theoretically, a temporal window with a length of $a=3$ is stacked in each iteration. After $N=5$ iterations, the size of the temporal window should be $a + (N-1) \times (a-1) = 11$. when we replace the $(i+j)$-th frame with ground-truth, if the$(i+j)$-th frame falls within the receptive field, the restoration of the $i$-th frame will benefit from the information obtained from that ground-truth frame.

Fig. \ref{img_recfield} displays the average restoration scores of video clips from two datasets when inserting ground-truth frames at different positions. It can be observed that when $j$ takes values from $1$ to $5$, the PSNR scores of the $i$-th frame all increase. This finding clearly demonstrates that CDUN can effectively leverage information from frames as far as $i+5$. This observation is consistent with the theoretical temporal receptive field length (from the $i-5$ frame to the $i+5$ frame). Notably, in dense degradations, such as haze and low-light conditions, all the pixels in the ground truth are discrepant significantly from the original frame. This discrepancy can severely interference with optical flow estimation, consequently impacting the experimental results. Consequently, this experiment is only conducted on sparse degradations, namely rain and snow.

\section{Conclusion}
This paper is the first study to successfully realize All-In-One video restoration. The proposed Cross-Consistent Deep Unfolding Network (CDUN) enables the restoration of different degraded videos with a single model. Specifically, CDUN consists of two core modules: (1) a well-designed SADE that can adaptively estimate the degradation matrices of differently degraded videos; and (2) a novel iterative optimization framework that can remove the degradation defined in the degradation matrices, from corrupted videos. By cascading the above two modules, CDUN achieves adaptive removal of diverse degradations. In addition, CDUN can capture complementary background information in adjacent frames by modeling inter-frame correlations within a temporal window, and can further expand the temporal receptive field by stacking windows. This mechanism enables the restoration of each frame to utilize the abundant features in long temporal-range frames. Consequently, by adaptive degradation removal and long-range feature utilization, the proposed CDUN is capable of effectively restoring videos corrupted by diverse degradations. Extensive experiments demonstrate that the proposed method achieves state-of-the-art performance in All-In-One VR.
\section{Limtation}
The proposed method is based on the generalized degradation model (i.e., Eq. (\ref{eq_allinone})). As a result, the proposed method is theoretically coupled with a degradation process in the form of Eq. (\ref{eq_allinone}). When applied to other degradation processes, the proposed method may have a potential risk of accuracy reduction. For example, motion blur is typically modeled as the convolution of a clean background with a blur kernel\cite{zhang2020deep,pan2019phase}, rather than being represented by a multiplicative and an additive factor as in Eq. (\ref{eq_allinone}).

%\ifCLASSOPTIONcompsoc
  %\section*{Acknowledgments}
%\else
  %\section*{Acknowledgment}
%\fi


\ifCLASSOPTIONcaptionsoff
  \newpage
\fi

\bibliographystyle{IEEEtran}
\bibliography{IEEEabrv,egbib}

%\begin{thebibliography}{1}

  %0.5em minus 0.4em\relax Harlow, England: Addison-Wesley, 1999.

%\end{thebibliography}

\begin{IEEEbiography}[{\includegraphics[width=1in,height=1.25in,clip,keepaspectratio]{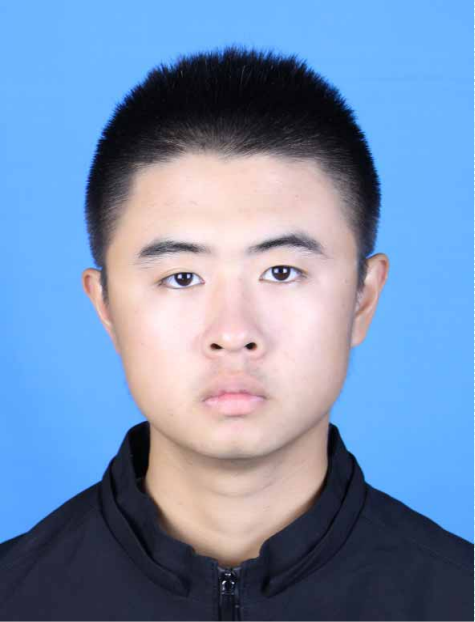}}]{Yuanshuo Cheng} received the B.Eng. degrees in College of Computer Science and Technology, China University of Petroleum, City Qingdao, China, in 2023. Now he is an M.S. at China University of Petroleum (East China), under the supervision of Prof. Shao. His current research interests include image restoration, computer vision, and deep learning.
\end{IEEEbiography}
\vspace{-30pt}
\begin{IEEEbiography}[{\includegraphics[width=1in,height=1.25in,clip,keepaspectratio]{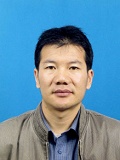}}]{Mingwen Shao} received his M.S. degree in mathematics from the Guangxi University, Guangxi, China, in 2002, and the Ph.D. degree in applied mathematics from Xi'an Jiaotong University, Xi'an, China, in 2005. He received the postdoctoral degree in control science and engineering from Tsinghua University in February 2008. Now he is a professor and doctoral supervisor at China University of Petroleum (East China). His research interests include machine learning, computer vision, and data mining.
\end{IEEEbiography}
\vspace{-30pt}
\begin{IEEEbiography}[{\includegraphics[width=1in,height=1.25in,clip,keepaspectratio]{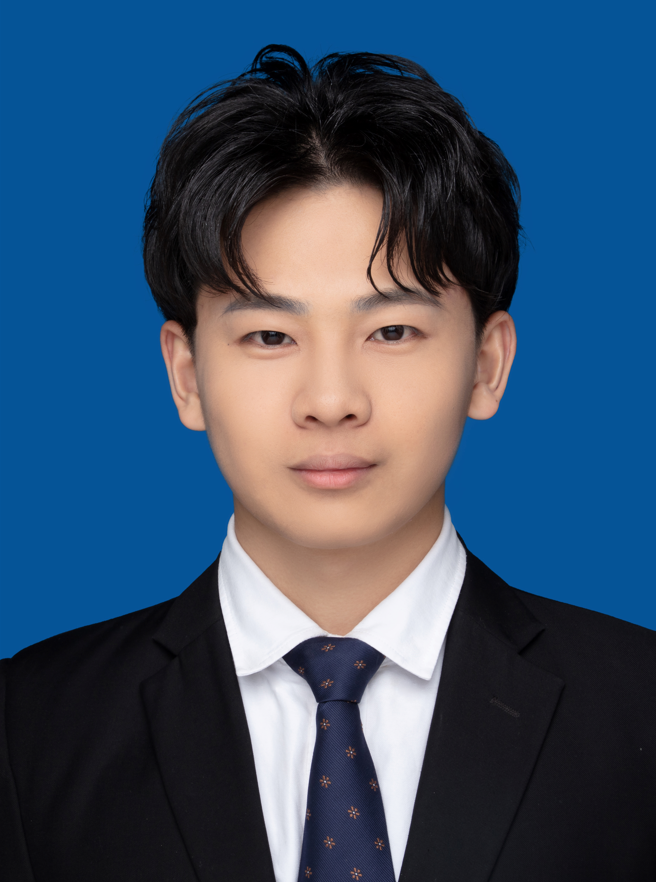}}]{Yecong Wan} received the B.Eng. degrees in College of Computer Science and Technology, China University of Petroleum, City Qingdao, China, in 2023. Now he is an M.S. at China University of Petroleum (East China), under the supervision of Prof. Shao. His current research interests include image restoration and computer vision.
\end{IEEEbiography}
\vspace{-30pt}
\begin{IEEEbiography}[{\includegraphics[width=1in,height=1.25in,clip,keepaspectratio]{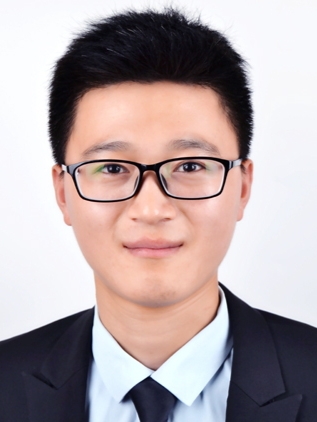}}]{Yuanjian Qiao} received the M.S. degree in electrical engineering and automation from the Qilu University of Technology (Shandong Academy of Sciences), Jinnan, China, in 2021. He is currently pursuing the Ph.D. degree under the supervision of Prof. M. Shao in the School of Computer Science and Technology, China University of Petroleum (East China), Qingdao, China. His current research interests include image restoration and deep learning
\end{IEEEbiography}
\vspace{-30pt}
\begin{IEEEbiography}[{\includegraphics[width=1in,height=1.25in,clip,keepaspectratio]{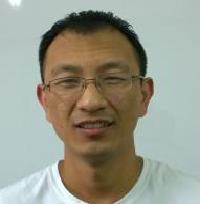}}]{Wangmeng Zuo} received the B.Sc., M.Sc., and Ph.D. degrees from the Harbin Institute of Technology, Harbin, China, in 1999, 2001, and 2007, respectively. He is currently a Professor with the School of Computer Science and Technology, Harbin Institute of Technology. His current research interests include computer vision, machine learning, and biometrics research.
\end{IEEEbiography}
\vspace{-30pt}
\begin{IEEEbiography}[{\includegraphics[width=1in,height=1.25in,clip,keepaspectratio]{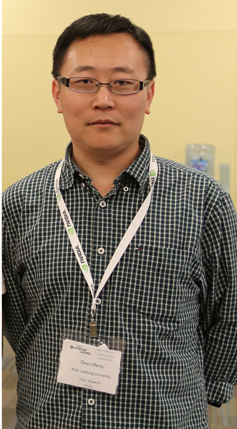}}]{Deyu Meng} received the B.Sc., M.Sc., and Ph.D. degrees from Xi’an Jiaotong University, Xi’an, China, in 2001, 2004, and 2008, respectively. He was a Visiting Scholar with Carnegie Mellon University, Pittsburgh, PA, USA, from 2012 to 2014. He is currently a Professor with the Institute for Information and System Sciences, Xi’an Jiaotong University. His current research interests include self-paced learning, noise modeling, and tensor sparsity.
\end{IEEEbiography}

%$\end{IEEEbiographynophoto}


%\vfill


\end{document}